\documentclass[journal]{IEEEtran}
\usepackage{amsmath}
\usepackage{amssymb}
\usepackage{times}
\usepackage{epsfig}
\usepackage{graphicx}
\usepackage{bm}
\usepackage{graphics}
\usepackage{subfigure}
\usepackage{epstopdf}
\usepackage{multirow}
\usepackage{color}
\usepackage{cite}
\ifCLASSINFOpdf
\else
\fi
\begin{document}
%
\title{Deep Ranking for Person Re-identification via\\ Joint Representation Learning}
%
%
%

\author{Shi-Zhe~Chen,
        Chun-Chao~Guo,~\IEEEmembership{Student~Member,~IEEE,}
        and~Jian-Huang~Lai,~\IEEEmembership{Senior~Member,~IEEE}
\thanks{This work was supported by the Natural Science Foundation of China (no. 61573387), the Guangdong Program (no. 2015B010105005) and the Guangzhou Program (no. 201508010032). \emph{(Corresponding author: Jian-Huang Lai.)}}
\thanks{S.-Z. Chen is with the School of Electronics and Information Technology, Sun Yat-sen University, Guangzhou 510006, China, and also with the Guangdong Province Key Laboratory of Information Security, China (e-mail: chenshizh@gmail.com).}
\thanks{C.-C. Guo and J.-H. Lai are with the School of Data and Computer Science, Sun Yat-sen University, Guangzhou 510006, China. C.-C. Guo is also with the Guangdong Provincial Key Laboratory of Digital Signal and Image Processing Techniques, China. J.-H. Lai is also with the Guangdong Province Key Laboratory of Information Security, China (e-mail: chunchaoguo@gmail.com; stsljh@mail.sysu.edu.cn).}
}

%
%

\markboth{IEEE TRANSACTIONS ON IMAGE PROCESSING}
{Shell \MakeLowercase{\textit{et al.}}: Bare Demo of IEEEtran.cls for Journals}

%



\maketitle

\begin{abstract}
This paper proposes a novel approach to person re-identification, a fundamental task in distributed multi-camera surveillance systems. Although a variety of powerful algorithms have been presented in the past few years, most of them usually focus on designing hand-crafted features and learning metrics either individually or sequentially. Different from previous works, we formulate a unified deep ranking framework that jointly tackles both of these key components to maximize their strengths.
 We start from the principle that the correct match of the probe image should be positioned in the top rank within the whole gallery set. An effective learning-to-rank algorithm is proposed to minimize the cost corresponding to the ranking disorders of the gallery. The ranking model is solved with a deep convolutional neural network (CNN) that builds the relation between input image pairs and their similarity scores through joint representation learning directly from raw image pixels. The proposed framework allows us to get rid of feature engineering and does not rely on any assumption.
An extensive comparative evaluation is given, demonstrating that our approach significantly outperforms all state-of-the-art approaches, including both traditional and CNN-based methods on the challenging VIPeR, CUHK-01 and CAVIAR4REID datasets. Additionally, our approach has better ability to generalize across datasets without fine-tuning.
\end{abstract}


%
\IEEEpeerreviewmaketitle

\section{Introduction}
\begin{figure}
\centering
\includegraphics[width=0.45\textwidth, angle=0]{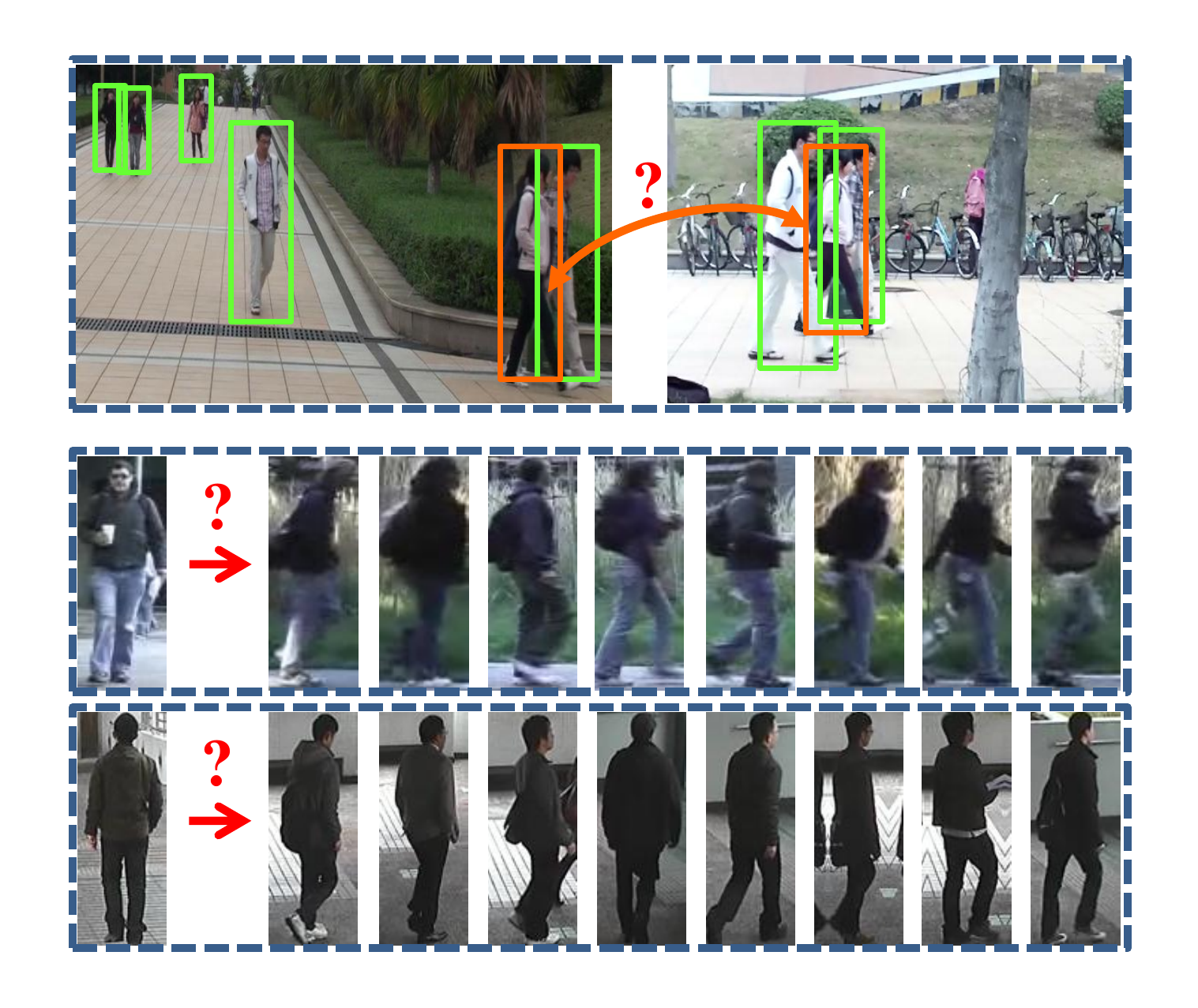}
\caption{Person re-identification remains challenging due to the drastic cross-view variations caused by illumination, occlusion, pose, etc. The images in the first row are taken from realistic surveillance systems, and those in the second and third rows come from the VIPeR and CUHK-01 datasets, respectively.}
\label{fig:intro}
\end{figure}

%
%
%
%
\IEEEPARstart{P}{erson} re-identification underpins many critical applications in long-term multi-camera tracking \cite{song2010tracking} and forensic search \cite{vezzani2013people}, and is increasingly receiving attention as a key component of video surveillance \cite{gong2014re}. Given an image of a target pedestrian captured by one camera, a person re-identification system attempts to recognize the occurrence of that target from a gallery of already-labeled subjects. Since the camera views of the realistic video surveillance system are usually disjoint, the system has to re-identify pedestrians based solely on visual cues most of the time. However, the appearance of a given individual undergoes drastic changes owing to complex variations in illumination, pose, viewpoint, occlusion, image resolution and camera setting, rendering person re-identification an unsolved and challenging problem (Figure \ref{fig:intro}).

Because the main difficulty in person re-identification arises from severe changes across non-overlapping camera views, an obvious solution is to design robust and discriminative descriptors for cross-view matching. Low-level features such as color (color histograms of different color spaces \cite{mignon2012pcca,xiong2014person,zheng2013reidentification,ma2014person}) and texture (LBP \cite{li2013locally}, Gabor \cite{gray2008viewpoint,li2013locally,zheng2013reidentification}) are commonly used for this purpose. Some studies have sought more distinct and reliable feature representation for pedestrians, including symmetry-driven accumulation \cite{farenzena2010person}, horizontal stripe-based partition \cite{liu2012person,zheng2013reidentification}, pyramid matching \cite{guo2014multi}, and salience matching \cite{zhao2013unsupervised,zhao2013person}. Unfortunately, it is extremely difficult to design a feature that is distinct, reliable and invariant to severe changes and misalignment across disjoint views.

Person re-identification has also been cast as a metric learning problem, resulting in significant performance improvements \cite{dikmen2011pedestrian,kostinger2012large,li2012human,mignon2012pcca,li2013learning,pedagadi2013local,xiong2014person,zheng2013reidentification}. These approaches typically extract hand-crafted features from the training set, and subsequently learn the metrics. From this perspective, metric learning essentially performs feature selection when learning the discriminative models. However, these approaches optimize two key components separately or sequentially. If feature representation is not reliable, some useful information is lost in the first step, and we cannot expect that the learnt metric obtained in the second step will have desirable performance. Hence, it will be a better choice to jointly learn feature representations and the metrics.

Different from those approaches, we propose a novel deep ranking algorithm for person re-identification. Instead of learning a metric over hand-crafted features, our approach learns joint representations and similarities for image pairs directly from the raw image pixels in a unified framework. Person re-identification can be cast as a retrieval problem: given one or more images of an unknown target, the re-identification task is to rank all individuals from the gallery according to their similarities to that target. We follow the principle to position the correct match of a probe at the top of the list against the gallery set. Hence, we penalize any violation of the ranking order by minimizing the cost corresponding to the sum of the rank of the true match of each probe. Leveraging the close connection between evaluation metrics for learning to rank and loss functions for classification \cite{yun2014ranking}, we formulate the person re-identification (ranking) task as a seemingly unrelated binary classification problem.

Inspired by its outstanding performance on numerous traditional computer vision tasks \cite{lecun1990handwritten,krizhevsky2012imagenet,chatfield2014return,zeiler2014visualizing,girshick2014rich,sun2014deep,chen2016DISC,toshev2014deeppose}, we utilize the deep convolutional neural network (CNN) to build the relation between a pair of pedestrian images and its similarity score. More specifically, our ranking model is built upon a CNN in which feature representation and metric learning are seamlessly integrated. Rather than employ the Euclidean or cosine distance between the features of a pair of images as the metric, we learn the joint representation of that pair and return a similarity score directly. At the training stage, we organize the labeled data into ranking units, each of which consists of a probe, its true match and corresponding reference set. Our deep network then learns a transformation that tends to assign the highest similarity score to the true match in each ranking unit.

Comprehensive evaluations and comparisons clearly demonstrate the marked superiority of our proposed approach over state-of-the-art person re-identification methods. To the best of our knowledge, there are currently two CNN-based person re-identification algorithms: Deep Metric Learning (DML) \cite{yi2014deep} and deep Filter Pairing Neural Network (FPNN) \cite{li2014deepreid}. Although our approach is not the first to address the person re-identification problem with deep learning, it is more suitable for re-identification and achieves better performance than either of them.

The contributions of this paper can be summarized as follows.
\begin{itemize}
  \item It proposes a unified deep ranking framework for person re-identification that directly predicts the similarity of a pair of pedestrian images via joint representation learning. There is no need to explicitly design feature representations, matching models or pre-processing. Our approach is more natural than previous re-identification approaches, including both traditional and deep learning based algorithms.
  \item An effective learning-to-rank algorithm is presented and integrated with the CNN. It penalizes ranking disorders in the gallery set, and tends to place the true match at the top.
  \item Extensive evaluation and analysis of the experimental results demonstrate the effectiveness of our approach. We carefully analyze each component of the framework for a fair self-evaluation, and further discuss key elements that may improve performance in a re-identification framework.
\end{itemize}

In the next section, we review the related work. We then present our proposed approach in Section \ref{sec:method}, followed by its optimization in Section \ref{sec:optimization}. Section \ref{sec:exp} presents an extensive comparison with state-of-the-art algorithms, and we analyze each component of our method. Section \ref{sec:conclusion} concludes the paper and discusses the future work.

\section{Related Work}
We review two streams of related work in terms of the technical components of this work: person re-identification and deep representation learning.

\subsection{Person Re-identification}
Many recent studies have addressed the person re-identification problem. Most of them focus primarily on either new descriptors or metric learning for person re-identification.

The aim of person re-identification descriptors is to generate discriminative signatures for pedestrians. Gray \emph{et al.} \cite{gray2008viewpoint} defined a feature space consisting of raw color channels in numerous color spaces and texture information captured by Gabor and Schmid filters, ensuring that Ensemble of Localized Features (ELFs) carrying more discriminative information were selected by boosting.
Tahir \emph{et al.} \cite{tahir2014cost} proposed a cost-and-performance-effective (CoPE) feature selection approach to identify both well-performing and cost-effective feature subset for person re-identification.
Faranzana \emph{et al.} \cite{farenzena2010person} proposed the Symmetry-Driven Accumulation of Local Features (SDALF) that exploited the symmetry property of a person through obtaining head, torso, and legs positions to handle view variations. Cheng \emph{et al.} \cite{cheng2011custom} extended the Pictorial Structure (PS) with their Custom Pictorial Structure (CPS) model to estimate body configurations and extract features from each body part. Ma \emph{et al.} \cite{ma2012bicov} developed the BiCov descriptor based on Gabor filters and the covariance descriptor to handle illumination variations. Kviatkovsky \emph{et al.} \cite{kviatkovsky2013color} developed an invariant intra-distribution structure of color under a wide range of imaging conditions. Yang \emph{et al.} \cite{yang2014salient} employed color naming and proposed the semantic Salient Color Names based Color Descriptor (SCNCD) that demonstrated robustness to photometric variance.
However, descriptors of visual appearance are highly susceptible to cross-view variations due to the inherent visual ambiguities and disparities caused by different view orientations, occlusions, illumination and background clutter. It is difficult to achieve a balance between discriminative power and robustness. In addition, some of these methods rely heavily on foreground segmentations, for instance, \cite{farenzena2010person} needs high-quality silhouette masks for symmetry-based partition, and \cite{kviatkovsky2013color} extracts color intra-distribution signatures only from foreground regions.

Metric learning approaches to re-identification usually follow a similar pipeline: extracting features for each image first, and then learning a metric with which the training data have strong inter-category differences and intra-category similarities. Prosser \emph{et al.} \cite{prosser2010person} developed an ensemble RankSVM to learn a subspace where the potential true match is given the highest ranking. Mignon \emph{et al.} \cite{mignon2012pcca} proposed the Pairwise Constrained Component Analysis (PCCA) to learn a projection into a low dimensional space in which the distance between pairs of data points respects the desired constraints, exhibiting good generalization properties in the presence of high dimensional data. In \cite{dikmen2011pedestrian}, a metric learning framework is used to obtain a robust Mahalanobis metric for Large Margin Nearest Neighbor classification with Rejection (LMNN-R). Zheng \emph{et al.} \cite{zheng2013reidentification} proposed the Relative Distance Comparison (RDC) approach to maximize the likelihood of a pair of true matches having a relatively smaller distance than that of a wrongly matched pair in a soft discriminant manner. In \cite{kostinger2012large}, the authors introduced the KISSME method from equivalence constraints based on a statistical inference perspective. Li \emph{et al.} \cite{li2013locally} partitioned the image spaces of two camera views into different configurations according to the similarity of cross-view transforms, and learned different metrics for different locally aligned common space. In \cite{li2013learning}, Li \emph{et al.} developed a Locally-Adaptive Decision Function (LADF) that jointly learned the distance metric and a locally adaptive thresholding rule. Pedagadi \emph{et al.} \cite{pedagadi2013local} utilized the Local Fisher Discriminant Analysis (LFDA) \cite{sugiyama2007dimensionality} to learn a subspace to reduce the dimensionality of the extracted high dimensional features. Xiong \emph{et al.} \cite{xiong2014person} further proposed and evaluated the performance of regularized PCCA (rPCCA), kernel LFDA (kLFDA) \cite{sugiyama2007dimensionality} and Marginal Fisher Analysis (MFA) \cite{yan2007graph} with different features and kernels. Other methods that deserve mentioning include salience matching \cite{zhao2013person} and mid-level filter learning \cite{zhao2014learning}.

The methods discussed in this subsection share three main drawbacks: (1)~their performance is largely limited by the representation power of the hand-crafted features; (2)~feature extraction and metric learning are considered as two independent components and optimized separately, so the interaction between them is not well explored; and (3)~the learnt metrics are fitted exclusively to the current scenario (dataset), and cannot be generalized to a new scenario without a significant deterioration in performance.

\subsection{Deep Learning}
Recently, approaches that extract features with deep learning structures, the deep CNNs in particular, have shown great potential in various computer vision tasks, including image classification \cite{krizhevsky2012imagenet}, object detection \cite{girshick2014rich}, face verification \cite{sun2014deep}, salient object detection \cite{chen2016DISC}, and pose estimation \cite{toshev2014deeppose}. Although deep learning for re-identification has not been fully investigated, the following works are close to our work in the spirit of learning image similarity or ranking. Hu \emph{et al.} \cite{hu2014discriminative} presented a new Discriminative Deep Metric Learning (DDML) method for face verification in the wild, and Wu \emph{et al.} \cite{wu2013online} employed deep learning architecture to learn a ranking model for image retrieval. However, they learned deep networks from hand-crafted features. Wang \emph{et al.} \cite{wang2014learning} proposed a deep ranking model with multi-scale CNNs to learn fine-grained image similarity directly from the image pixels.

To our knowledge, two deep learning based person re-identification algorithms have been proposed. Yi \emph{et al.} \cite{yi2014deep} utilized a Siamese CNN with a symmetry structure comprising two sub-nets connected by a cosine layer, and proposed a DML approach for re-identification. Given a pair of images, the deep network extracts features of each image independently, and then uses their cosine distance as the metric. Li \emph{et al.} \cite{li2014deepreid} designed an FPNN that takes two images of pedestrians as input and determines whether they have the same identity. The notable difference between these two algorithms is that the FPNN learns the joint representation of two images, while the DML does not. However, learning a network for binary classification does not seem to be a good choice, because positive pairs are much fewer than negative pairs, and thus the learned network tends to predict most input pairs as negative ones due to the great imbalance of training data \cite{he2009learning}.

To address these problems, we propose a unified deep learning-to-rank framework that learns joint representation and similarities of image pairs directly from image pixels.

\section{Deep Ranking Framework} \label{sec:method}

\subsection{Overview}
\begin{figure*}
\centering
\includegraphics[width=1\textwidth, angle=0]{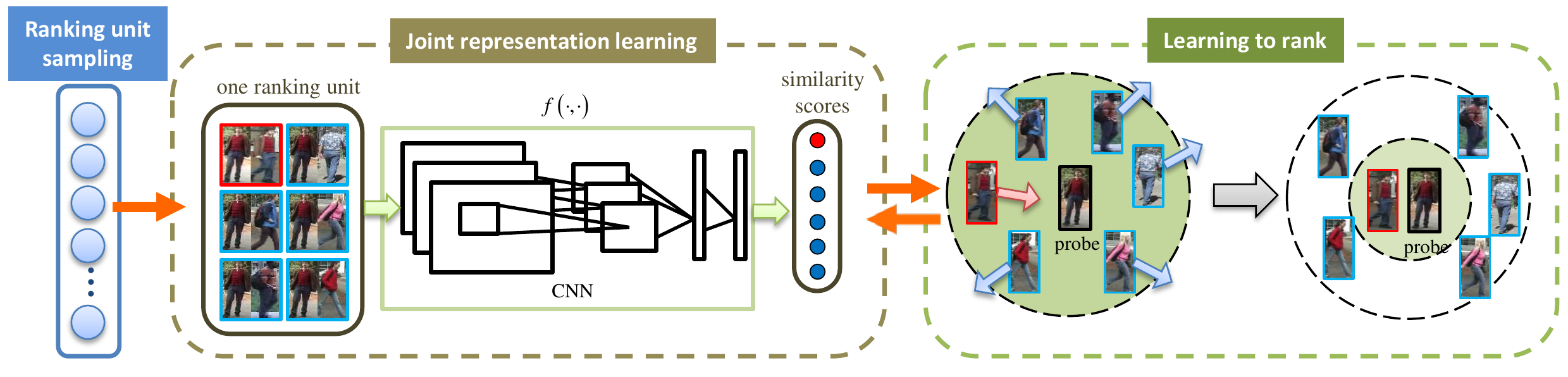}
\caption{Illustration of our proposed deep ranking framework, which comprises two key components: deep joint representation learning and learning to rank. We aim to learn a deep CNN that assigns a higher similarity score to the positive pair (marked in red) than any negative pairs (marked in blue) in each ranking unit. Best viewed in color.}
\label{fig:overview}
\end{figure*}

In this section, we describe the proposed approach in detail. Figure \ref{fig:overview} gives an illustration of our proposed framework. At the training stage, the labeled data are organized into ranking units and then fed into the deep CNN. The CNN is utilized to model the transformation $f\left ( \cdot ,\cdot  \right )$ from a pair of pedestrian images to its similarity score. Since the correct match should be positioned at the top of the gallery, we penalize ranking disorders by minimizing the sum of the ranks of positive pairs in each ranking unit. We formulate these two components into our deep ranking framework and perform joint optimization. The learnt CNN conducts similarity computing in one shot at the test time.

\subsection{Formulation}
Before delving deeper into the formulation, we describe some of the terminologies associated with our problem that will be used later. Without loss of generality, let us consider solving the following person re-identification problem in a single-shot case for convenience. Suppose that we are given a training set $\mathcal{X}=\left \{ \left ( x_{i}^{A},x_{i}^{B} \right )|i=1,2,...,N \right \}$, where $\left ( x_{i}^{A},x_{i}^{B} \right )$ is a pair of images of the \emph{i-th} person captured by cameras \emph{A} and \emph{B}, respectively, and $N$ is the number of pedestrians. For a probe image $x$ to be matched against gallery set $\mathcal{G}$, a ranking list should be generated according to the similarity between $x$ and each image in $\mathcal{G}$. There exists only one correct match $x^+$, which should be placed in the top rank by the learnt ranking model. All other samples in the gallery space are considered to be negative matches, denoted by $\mathcal{G^-}$.

Intuitively, if the learnt ranking model is perfect, the correctly matched pair will be assigned a higher similarity score than a mismatched one, which can be expressed as
\begin{equation}f\left( x,x^+ \right )>f\left ( x,y\right ),\forall y\in \mathcal{G^-},\end{equation}
where $f\left ( \cdot ,\cdot  \right ):\mathcal{X}\times\mathcal{X}\mapsto\mathbb{R} $ is the learnt similarity metric for an image pair. The rank of $x$ with respect to $\mathcal{G^-}$ can be expressed as a sum of the 0-1 loss function:
\begin{equation}\textrm{rank}\left ( x|\mathcal{G^-} \right )=\sum_{y\in \mathcal{G^-}}{I\left \{ f\left ( x,x^+ \right )-f\left ( x,y \right )<0 \right \}},\end{equation}
where $I\left ( \cdot  \right )$ is an indicator function whose value is 1 when the expression is true, and 0 otherwise. We propose our learning-to-rank framework based on two main considerations. First, our aim is to place the true match, $x^+$, at the top with regard to $\mathcal{G^-}$. In other words, $\textrm{rank}\left ( x|\mathcal{G^-} \right )$ has to be small. Second, for two mismatches, $y_i,y_j\in\mathcal{G^-}$, we have no idea which is more similar to a given probe $x$, and simply ignore the intra-ranking orders of $\mathcal{G^-}$. Therefore, we formulate the objective function as follows
\begin{equation}
\begin{aligned}
J&=\sum \limits_{x}{\textrm{rank}\left ( x|\mathcal{G^-} \right )} \\
&=\sum \limits_{x}{\sum \limits_{y\in \mathcal{G^-}}{I\left \{ f\left ( x,x^+ \right )-f\left ( x,y \right )<0 \right \}}}.
\end{aligned}
\end{equation}
This formulation minimizes the cost corresponding to the sum of the gallery ranking disorders of each probe.

Unfortunately, dealing directly with the 0-1 loss function leads to a non-differentiable optimization problem. The most common solution to this problem is to upper-bound the 0-1 loss by an easy-to-optimize function. Inspired by \cite{yun2014ranking}, we utilize logistic loss function $\sigma \left ( x \right )=\log_2\left ( 1+2^{-x}\right )$ to replace $I\left\{x<0\right\}$, and rewrite the objective function as
\begin{equation}J=\sum_{x}{\sum_{y\in \mathcal{G^-}}{\sigma \left ( f\left ( x,x^+ \right )-f\left ( x,y \right ) \right )}}.\end{equation}

In this model, the most critical component is learning similarity metric $f\left ( \cdot ,\cdot  \right )$. Conventional methods usually design hand-crafted features and subsequently learn a Mahalanobis metric to maximize the inter-class variations and minimize the intra-class variations. In this work, we propose to take advantage of deep CNNs to learn $f\left(\cdot,\cdot\right)$ directly from raw image pixels rather than from hand-crafted features. In the following subsection, we introduce the deep network architecture used in our ranking framework.

\subsection{Network Architecture}
\begin{figure*}
\centering
\includegraphics[width=1\textwidth, angle=0]{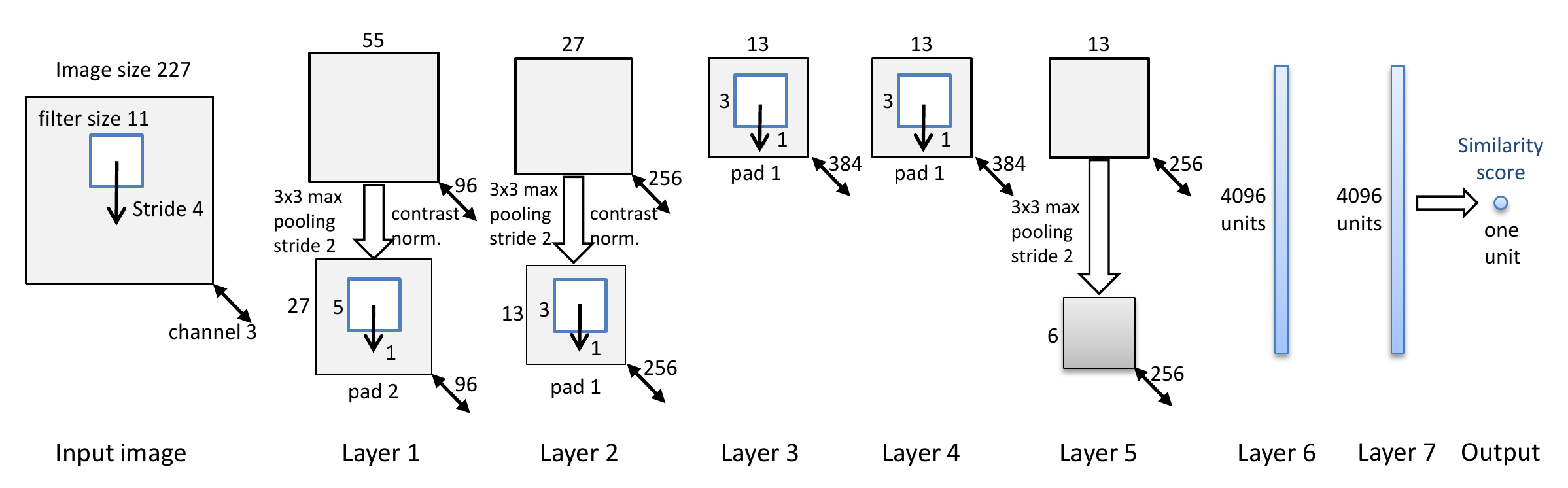}
\caption{Architecture of our deep network. A pair of three-channel pedestrian images is first stitched, and then a $227\times227$ random crop is presented as the input, which is convolved with 96 different first layer filters, each of size $11\times11$, using a stride of 4 in both $x$ and $y$. The resulting feature maps are then passed through a rectified linear unit (ReLU; not shown in this figure), max-pooled ($3\times3$ regions with stride 2), and contrast normalized across the feature maps to give 96 different $27\times27$ feature maps. Similar operations are repeated in the second to fifth layers. The last three layers are fully connected, taking features from the top convolutional layer as the input in vector form. Finally, a similarity score for the pair is returned.}
\label{fig:net_arch}
\end{figure*}
Our deep network learns mapping $f(\cdot,\cdot)$ from two images to their similarity score directly. It comprises five convolutional layers to extract features hierarchically, followed by three fully connected layers. Figure \ref{fig:net_arch} shows the detailed structure of our network, which is similar to the popular \emph{AlexNet} \cite{krizhevsky2012imagenet}. We propose to learn joint representation for an image pair (explained in Section \ref{sssec:joint}). Therefore, a notable difference is that we simply stitch two pedestrian images horizontally to form an image that is used as input. More specifically, the images in the pair are both resized to $H\times W$ (here, $H=2W$, and we set $H=256$ in the experiments) and then stitched together to form a square image for input. This approach ensures that the aspect ratio of the images remains nearly unchanged, and we do not need to design a new network architecture with two entrances.
The convolution operation is expressed as
\begin{equation}
\mathbf{x}^{(l)}_{i}=\textrm{relu}\left( b_{i}^{(l)}+ \sum_{j}{\mathbf{k}_{ij}^{(l)}\otimes\mathbf{x}_{j}^{(l-1)}}\right),
\end{equation}
where $\mathbf{x}^{(l)}_{i}$ and $\mathbf{x}^{(l-1)}_{j}$ denote the $i$-th output channel at the $l$-th layer and the $j$-th input channel at the $(l{-}1)$-th layer, respectively; $\mathbf{k}_{ij}^{(l)}$ is the convolutional kernel between the $i$-th and $j$-th feature map; and $b_{i}^{(l)}$ is the bias of the $i$-th map. The Rectified Linear Unit (ReLU) is used as the neuron activation function, denoted as $\textrm{relu}(x)=\max(x,0)$. Max-pooling is utilized in the first, second and fifth convolutional layers, formulated as
\begin{equation}
\mathbf{x}^{(l)}_{(i,j)}=\max_{\forall(p,q)\in\Omega_{(i,j)}}{\mathbf{x}^{(l)}_{(p,q)}},
\end{equation}
where $\Omega_{(i,j)}$ stands for the pooling region with index $(i,j)$. Since the variations in poses and viewpoints always appear, max-pooling enhances the robustness to small translations \cite{boureau2010theoretical}. Max-pooling at the first two layers is followed by local response normalization, leading to feature maps that are robust to illumination and contrast variations.

The last three layers are fully connected, expressed as
\begin{equation}
\mathbf{x}^{(l)}=\mathbf{w}^{(l)}\cdot \mathbf{x}^{(l-1)} + \mathbf{b}^{(l)},
\end{equation}
where $\mathbf{w}^{(l)}$ and $\mathbf{b}^{(l)}$ are the weight and bias, respectively.
The first two fully connected layers reduce the dimensionality of the extracted joint features from 9216 ($6\times6\times256$) to 4096, and form highly compact and predictive features, denoted by $\phi(x,y)$. The last layer acts as the similarity/distance metric for $\phi(x,y)$, which can be expressed as
\begin{equation}
f(x,y)=\left \langle \phi(x,y),\mathbf{w} \right \rangle+b,
\end{equation}
where $f\left ( \cdot ,\cdot  \right )$ denotes the similarity metric as before, and $\left \langle \cdot,\cdot \right\rangle $ denotes the inner-product between the two vectors. Our network is capable of jointly learning the features and similarity metric with supervised similarity information provided by the proposed ranking algorithm, which characterizes the relative similarity ranking orders.

Note that the activation function for all layers (except the last layer) is the ReLU. Dropout \cite{krizhevsky2012imagenet} is used in the first two fully connected layers to alleviate over-fitting.

Our reasoning for employing this very deep network architecture for person re-identification is as follows. Since the appearance of a given pedestrian undergoes drastic changes due to complex variations in illumination, pose, viewpoint, camera setting and background clutter across camera views, we argue that the network should be deep enough to handle the inherent visual ambiguities. Further, deeper network learning requires more training samples, but we are given only small-scale labeled data, particularly in a single-shot modality. For instance, the well-known VIPeR dataset \cite{gray2008viewpoint}, which has only 1,264 images for 632 subjects, is far from sufficient for network learning. It seems that the depth of the network is necessarily limited by the amount of training data. In the experiment, we show that this dilemma can be resolved by pre-training and other strategies, as explained in Section \ref{sec:exp}.

\section{Optimization} \label{sec:optimization}
Our network is trained using the stochastic gradient descent (SGD) algorithm with momentum. Training data $\mathcal{X}$ are organized into mini-batches consisting of several ranking units. The training errors are computed for each mini-batch, and back-propagated to the lower layers.
\subsection{Ranking Unit Sampling}
\begin{figure}
  \centering
  \includegraphics[width=0.35\textwidth, angle=0]{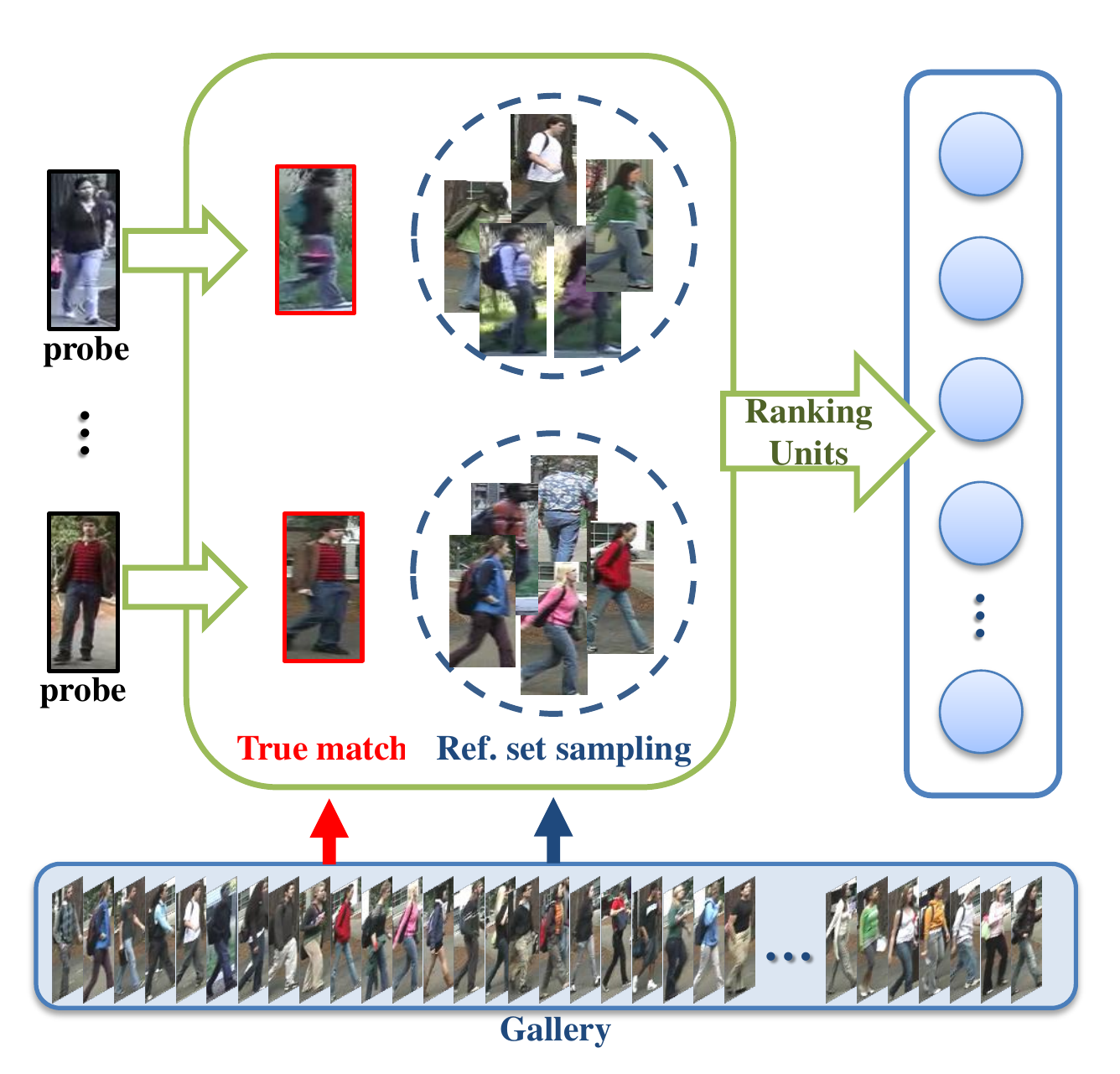}
  \caption{Illustration of ranking unit sampling.}
  \label{fig:ranking_unit}
\end{figure}
As discussed in the previous section, we organize the training data into ranking units. Here, we take only a subset of $\mathcal{G^-}$, i.e., $\mathcal{R}_x\subseteq \mathcal{G^-}$, into account. It is considered the reference set for probe image $x$. Each unit comprises a probe $x$, its true match $x^+$, and its corresponding reference set $\mathcal{R}_x$. Note that the reference set $\mathcal{R}_x$ is randomly sampled given probe $x$ (Figure \ref{fig:ranking_unit}). We consider $\mathcal{R}_x$ alone rather than the full $\mathcal{G^-}$ for three reasons: (1)~there is no need to load all data into memory at the training stage if we simply sample a subset of $\mathcal{G^-}$, making it more practical for large-scale learning; (2)~since reference set $\mathcal{R}_x$ is randomly sampled, the training data of each batch possess a high degree of diversity, which is of great importance to learning; and (3)~dealing with a random subset $\mathcal{R}_{x}$ in each iteration is approximately equivalent to taking the whole of $\mathcal{G^-}$ into account with sufficient iteration times (as explained in Section \ref{sssec:ranking_unit}). Therefore, the loss is formulated as
\begin{equation}\mathcal{L}=\sum_{x}{\sum_{y\in \mathcal{R}_{x}}{\sigma \left ( f\left ( x,x^+ \right )-f\left ( x,y \right ) \right )}}.\end{equation}
The gradients of the ranking loss with respect to the similarities within a ranking unit are

\begin{equation} \frac{\partial\mathcal{L}}{\partial f(x,x^{\prime})}=\left\{
\begin{aligned}
&\frac{\delta(x^{\prime},x^+|x)}{1+\delta(x^{\prime},x^+|x)}& &{x^{\prime}\in\mathcal{R}_{x}}\\
&\sum_{y\in \mathcal{R}_{x}}{\frac{-\delta(y,x^+|x)}{1+\delta(y,x^+|x)}}& &{x^{\prime}=x^+}
\end{aligned}
\right.
\end{equation}
where $\delta(i,j|x)=2^{f(i,x)-f(j,x)}$. The back-propagation algorithm adjusts $f(\cdot,\cdot)$ such that $f(x,x^+)$ is assigned the highest similarity score in the corresponding ranking unit.

Let $\mathcal{X}^A$ and $\mathcal{X}^B$ denote the training data captured by cameras \emph{A} and \emph{B}, respectively. For a probe image from one camera (say, camera \emph{A}), $x_{i}^{A}$, the correct match should be $x_{i}^{B}$. Here, the reference set can be expressed as $\mathcal{R}_{x_{i}^{A}}\subseteq \mathcal{G}^{-}_{x_{i}^{A}}$, where $\mathcal{G}^{-}_{x_{i}^{A}}=\left \{ x_{j}^{B}|x_{j}^{B}\in \mathcal{X}^B,j\neq i \right \}$. When training our deep network, we set $|\mathcal{R}_x|=1$ at the beginning, where $|\cdot |$ is the cardinality of a set. Note that the ranking unit now degrades into a simple triplet constraint, i.e., $f(x,x^+)>f(x,x^-)$ for a triplet $t=(x,x^+,x^-)$. As the learning procedure progresses, we gradually increase the cardinality of the reference set $|R_x|$ in each mini-batch up to 4. From another perspective, the positive pairs and the same number of negative pairs are fed into the deep network at the beginning for balancing. It is not a complicated task to satisfy the triplet constraints tentatively. Since there are much more negative pairs than positive ones, we gradually increase the number of negative samples up to a ratio of 4:1. Now, the problem becomes increasingly difficult because we want the ranking model to position the correct match at the top against a reference set that is increasing in size. In this way, a discriminative ranking model is obtained.
\subsection{Training Strategies}
Several critical training strategies are discussed in this subsection.

\noindent \textbf{Pre-training} - As previously discussed, a large amount of training data is needed for learning because of the great depth of our network. Since the available labeled data at hand are scarce, we use the labeled data collected from other scenarios (datasets) to assist network learning even though they are subject to quite different distributions. In our experiment, we first use large-scale data to learn a pre-trained model. For each specific scenario (dataset), we initialize the parameters with that pre-trained model, and then fine-tune all layers by back-propagation through the whole network with the given training set. We find experimentally that pre-training is a critical component that boosts performance significantly.

\noindent \textbf{Relaxing the cross-view constraint} - Recall that the reference set $\mathcal{R}_{x_{i}^{A}}$ for probe image $x_{i}^{A}$ is rigorously sampled from $\mathcal{G}^{-}_{x_{i}^{A}}=\left \{ x_{j}^{B}|x_{j}^{B}\in \mathcal{X}^B,j\neq i \right \}$. In a single-shot modality, the data are insufficient to construct a reference set exhibiting strong diversity. Intuitively, distinguishing two persons from the same camera is relatively easy, and it also helps to learn the similarity metric. Under this consideration we relax the cross-view constraint and sample the reference set from both camera views, i.e., $\mathcal{G}^{-}_{x_{i}^{A}}=\left \{ x_{j}|x_{j}\in \mathcal{X},j\neq i \right \}$.

\noindent \textbf{Data augmentation} - It is a common method to artificially enlarge the training data using label-preserving transformations to reduce over-fitting \cite{krizhevsky2012imagenet,chatfield2014return}. We also employ similar data augmentation in the form of random crops and horizontal flips. The only notable difference is that the flips are not generated by flipping the images around the $y$-axis directly. Since our input is essentially a pair of images, and we thus flip the image in a different way. Each sub-image is flipped around its own central vertical axis with probability 0.5, and the two sub-images further exchange their positions, also with probability 0.5, which increases the size of our training set by a factor of 8. We perform random crops by randomly extracting $227\times227$ patches from the original images (or their horizontal reflections), and then train our network on the extracted patches. At the test time, we deterministically extracted the central crop of the image in addition to its horizontal reflections, and returned the average of these eight scores. We also tried to extract five patches, including the center and four corner patches, as in \cite{krizhevsky2012imagenet}, but achieved similar results.

\section{Experiment} \label{sec:exp}
In this section, we report the results of extensive experiments carried out to compare our approach with state-of-the-art approaches, including both traditional and deep learning based methods, and to evaluate each component of our method in detail. Although the superiority of our approach comes from the framework as a whole, we also carefully assessed each component to give a fair self-evaluation.

\begin{figure}
  \centering
  \subfigure[Samples from the VIPeR dataset \cite{gray2008viewpoint}]{
  \label{subfig:cmc_viper}
  \includegraphics[width=0.38\textwidth, angle=0]{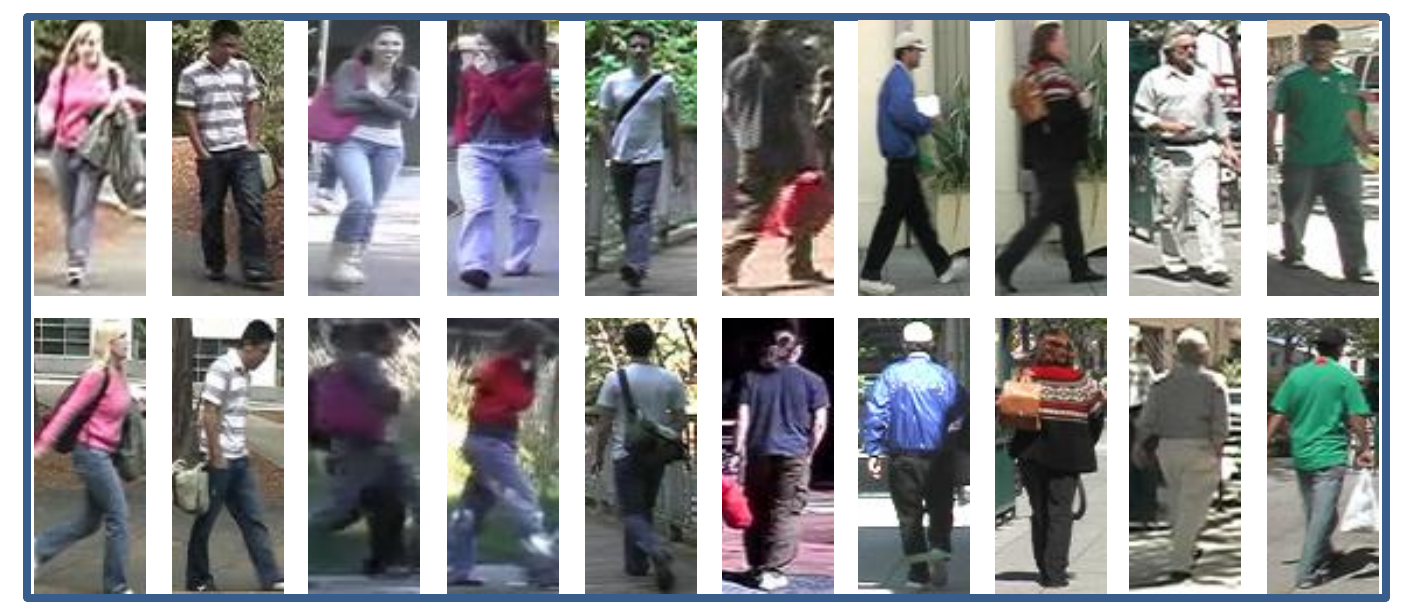}}
  \subfigure[Samples from the CUHK-01 dataset \cite{li2012human}]{
  \label{subfig:cmc_cuhk}
  \includegraphics[width=0.38\textwidth, angle=0]{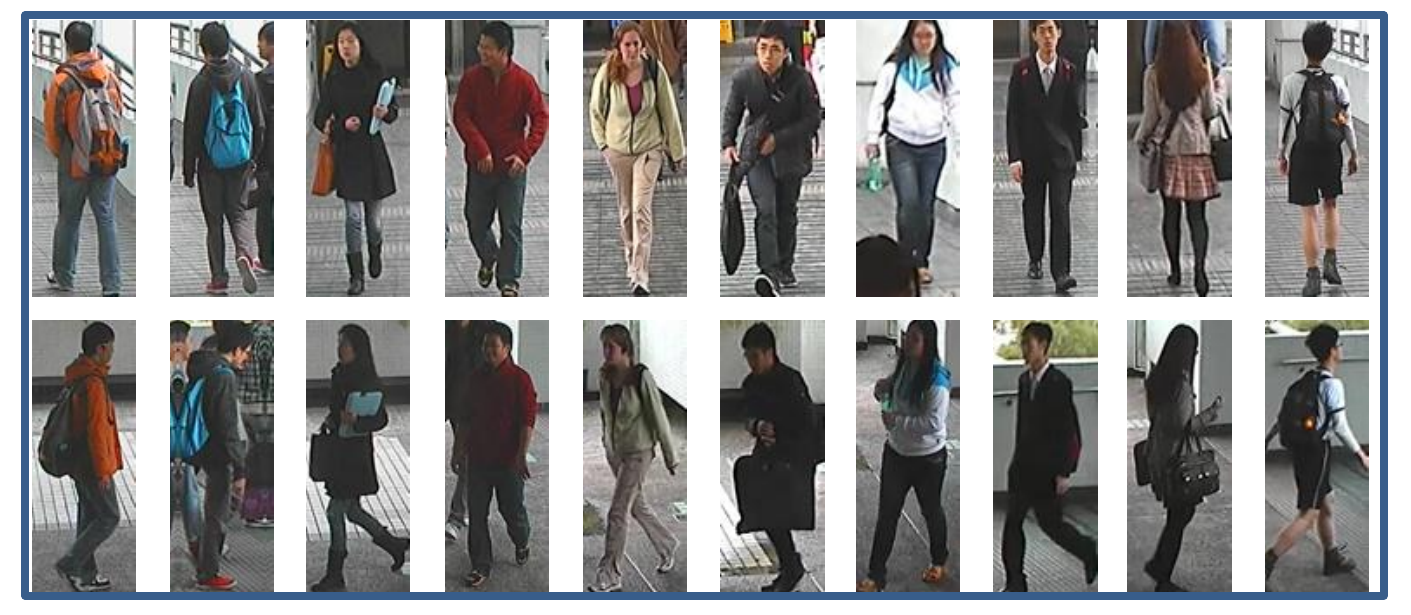}}

  \subfigure[Samples from the CAVIAR4REID dataset \cite{cheng2011custom}]{
  \label{subfig:cmc_caviar}
  \includegraphics[width=0.38\textwidth, angle=0]{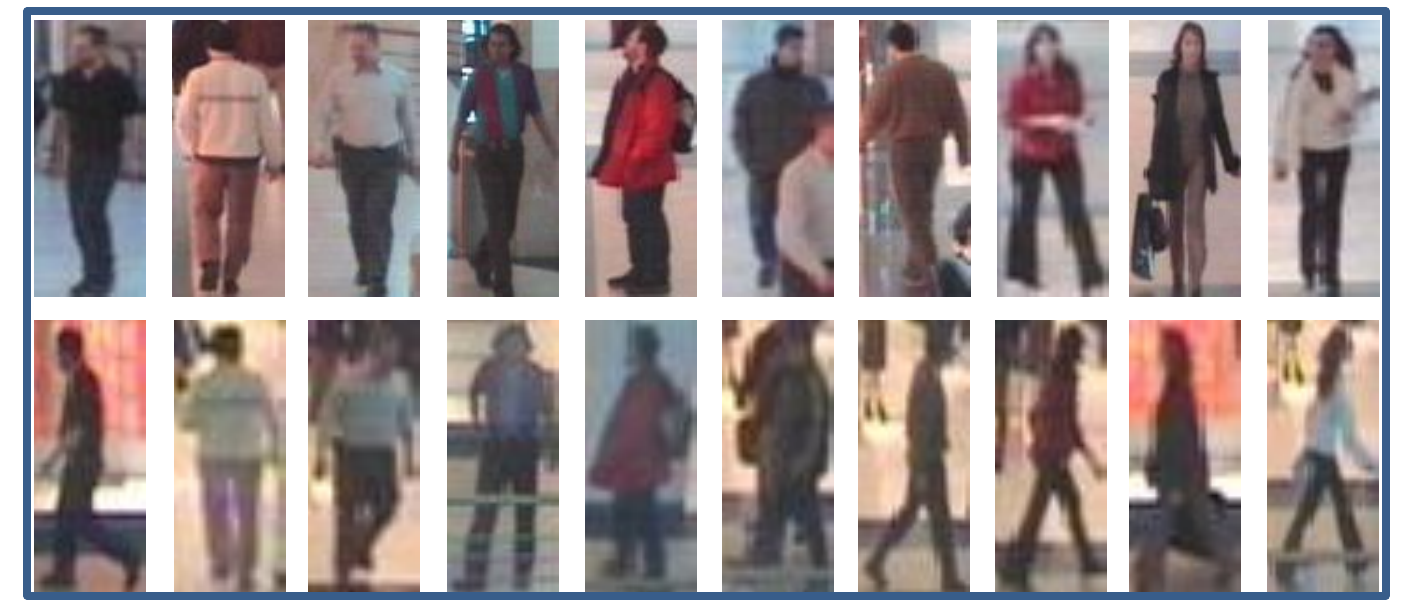}}
  \caption{Samples of pedestrian images observed in different camera views. Each column corresponds to the same identity.}\label{fig:samples}
\end{figure}

\subsection{Experimental Settings}
\noindent \textbf{Datasets} - To validate our approach, we performed experiments on three benchmark datasets: the VIPeR dataset \cite{gray2008viewpoint}, the CUHK-01 dataset \cite{li2012human}, and the CAVIAR4REID dataset \cite{cheng2011custom}. These datasets are highly challenging because of the different camera settings, geometric deformations and photometric variations in different views (Figure \ref{fig:samples}).

\noindent \textbf{Evaluation protocol} - We adopted the single-shot modality on the VIPeR and CUHK-01 datasets, and multi-shot modality (with both $N=5$ and $N=10$) on the CAVIAR4REID dataset as most previous studies did to allow extensive comparison. Following the commonly used evaluation protocol in \cite{gray2008viewpoint}, we randomly partitioned the dataset into two parts, one half for training and the other for testing, without any overlap in person identities. Each probe image was matched against the gallery set, and the rank of the true match was obtained. The rank-\emph{k} recognition rate is the expectation of the matches at rank \emph{k}, and the cumulative values of the recognition rate at all ranks were recorded as the one-trial Cumulative Matching Characteristic (CMC) result. We repeated the evaluation ten times, and here report the average CMC curve to achieve stable statistics. Note that the CUHK-01 dataset has more than one image of each person, and thus we randomly select one to form the gallery. For the CAVIAR4REID dataset, we used all ten images per view for multi-shot setting with $N=10$, and randomly select five images for the setting with $N=5$.

\noindent \textbf{Implementation details} - In our experiment, the CUHK-02 dataset \cite{li2013locally} was used to learn a pre-trained model. Note that the CUHK-02 dataset contains five pairs of views (P1-P5), and P1 is also called the CUHK-01 dataset. The samples from P1 (i.e., the CUHK-01 dataset) were excluded when learning the pre-trained model, because the CUHK-01 was used for evaluation. This ensures that no sample from the test set was used during the pre-training stage by mistake. We implemented our model under the open source Caffe CNN library \cite{jia2014caffe}, and trained it using the SGD with momentum of 0.9, weight decay of 0.0005, and learning rates of $10^{-4}$. The parameters were initialized with the model in \cite{krizhevsky2012imagenet}. We fixed the mini-batch size of 16 ranking units. The positive-negative ratio is set as 1:1 initially. We gradually increased the size of reference set $\mathcal{R}_x$ to 2 and 4, whilst the ratio changes to 1:2 and 1:4 accordingly (see Figure \ref{fig:ranking_unit}).

\subsection{Comparison with State-of-the-art Methods} \label{ssec:cmp}
\begin{figure*}
  \centering
  \subfigure[VIPeR dataset]{
  \label{subfig:cmc_viper}
  \includegraphics[width=0.32\textwidth, angle=0]{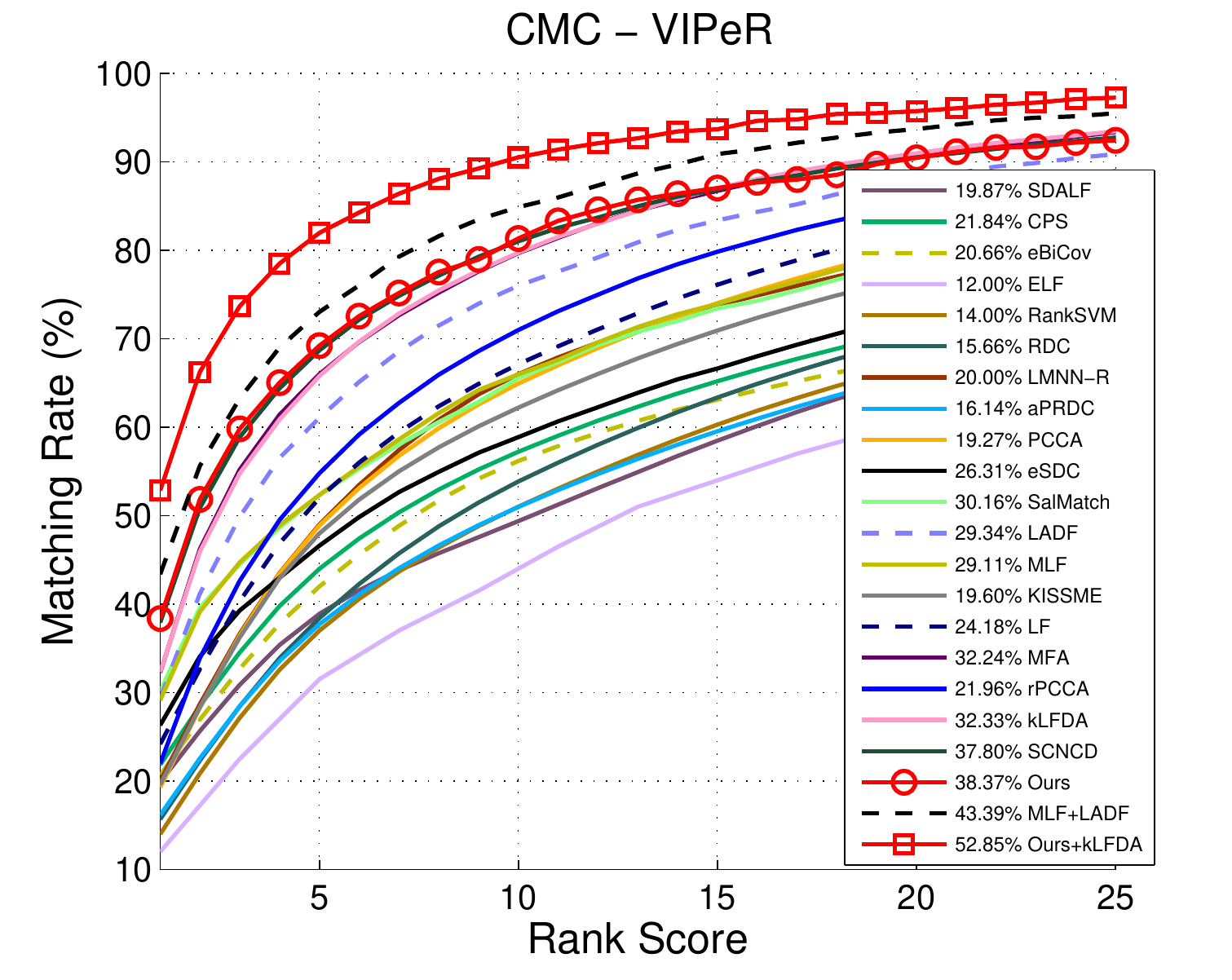}}
  \subfigure[CUHK-01 dataset]{
  \label{subfig:cmc_cuhk}
  \includegraphics[width=0.32\textwidth, angle=0]{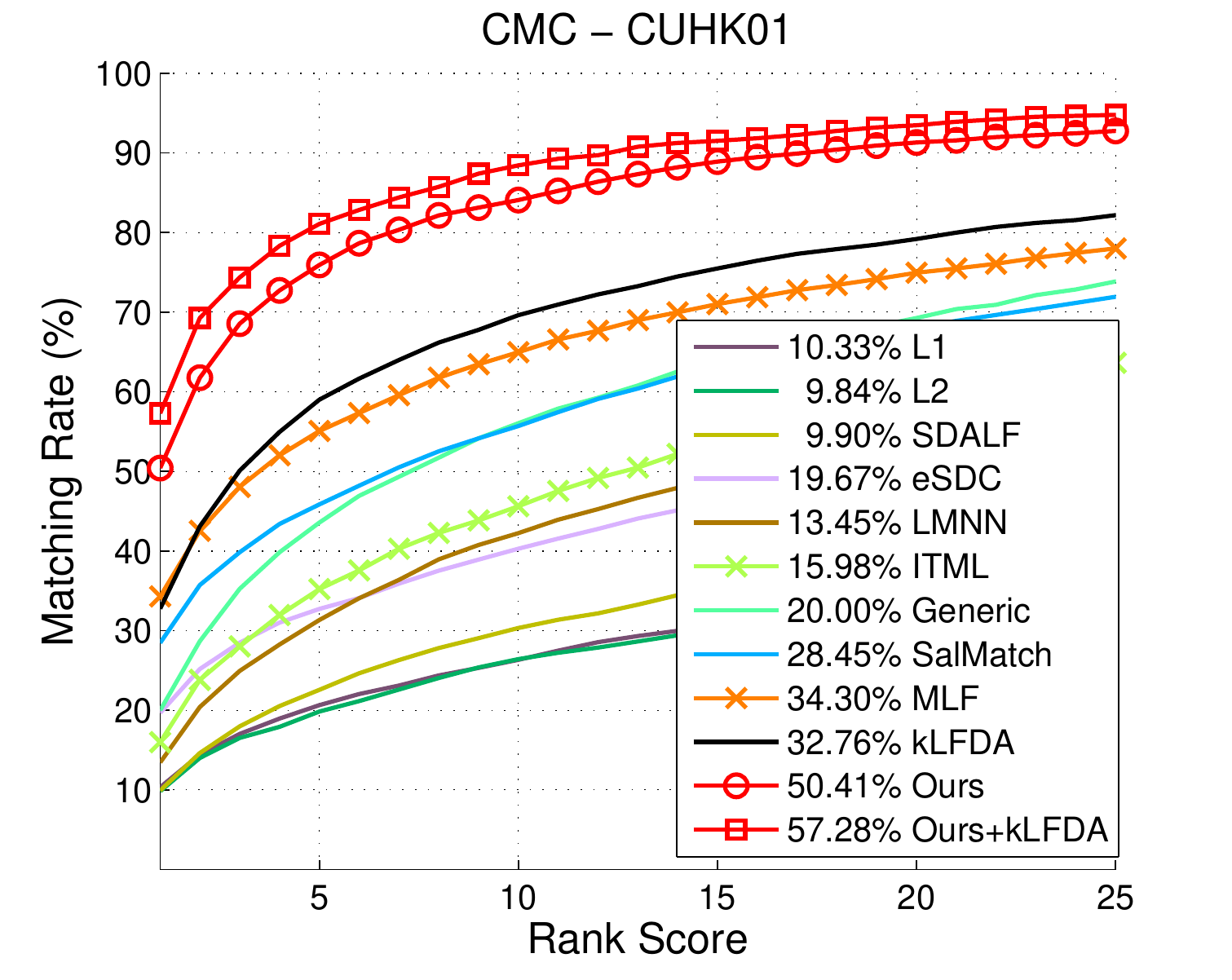}}
  \subfigure[CAVIAR4REID dataset]{
  \label{subfig:cmc_caviar}
  \includegraphics[width=0.32\textwidth, angle=0]{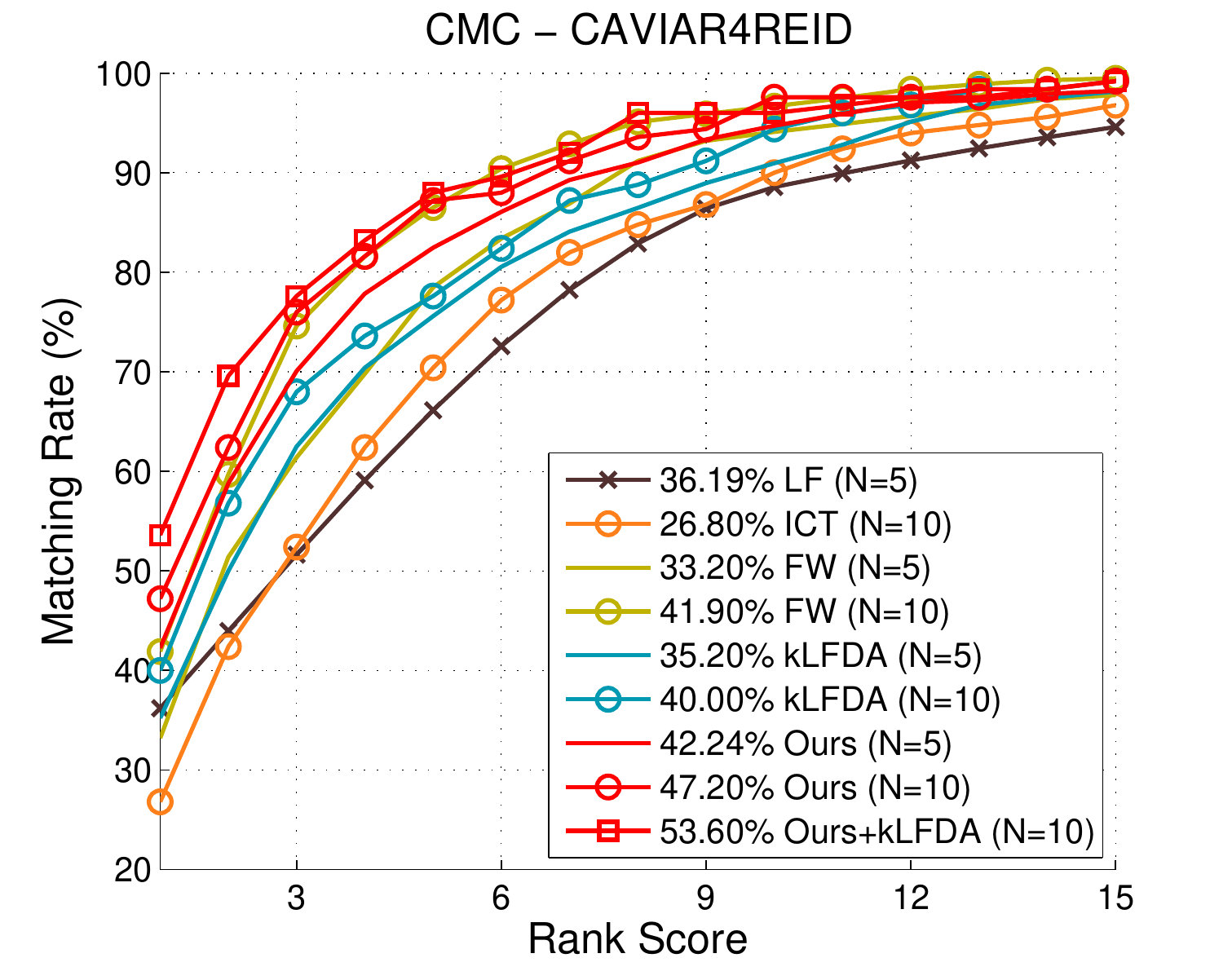}}
  \caption{Performance comparisons with state-of-the-art approaches using CMC curves on the VIPeR ($p=316$), CUHK-01 ($p=486$) and CAVIAR4REID ($p=25$) datasets. In the legends, we also report the rank-1 matching rate for each approach. Best viewed in color.}\label{fig:cmc}
\end{figure*}

In this section, we compare our proposed method with the following state-of-the-art approaches: ELF \cite{gray2008viewpoint}, SDALF \cite{farenzena2010person}, LMNN \cite{weinberger2009distance}, ITML \cite{davis2007information}, eSDC \cite{zhao2013unsupervised}, Generic Metric \cite{li2012human}, Salience Matching (SalMatch) \cite{zhao2013person}, Mid-Level Filter (MLF) \cite{zhao2014learning}, eBiCov \cite{ma2012bicov}, PCCA \cite{mignon2012pcca}, LF \cite{pedagadi2013local}, LADF \cite{li2013learning}, MFA \cite{xiong2014person}, kLFDA \cite{xiong2014person}, rPCCA \cite{xiong2014person}, RDC \cite{zheng2013reidentification}, attribute-based PRDC (aPRDC) \cite{liu2012person}, CPS \cite{cheng2011custom}, RankSVM \cite{prosser2010person}, LMNN-R \cite{dikmen2011pedestrian}, KISSME \cite{kostinger2012large}, SCNCD \cite{yang2014salient}, ICT \cite{avraham2012learning}, and Feature Warps (FW) \cite{martinel2015re}. Note that not all of these approaches have reported results for all three datasets. For instance, VIPeR is the most widely used benchmark, and thus several researchers have reported the results of various approaches on VIPeR but not CUHK-01 or CAVIAR4REID. Here, we compare our method with the foregoing methods if available.

\subsubsection{Performance on the VIPeR dataset}
The VIPeR dataset \cite{gray2008viewpoint} contains 632 person image pairs. Each pair has two images of the same person observed from different views, resized to $128\times48$. Most of the approaches considered have CMC curves reported for this dataset, hence we give a more detailed comparison for VIPeR. Figure \ref{subfig:cmc_viper} shows the CMC curves up to rank 25 comparing our method with state-of-the-art methods. It is obvious that our method gives the best result in the main. To present the quantized comparison results more clearly, we also summarize the performance comparison at several top ranks in Table \ref{tab:cmc_viper}. It can be seen that, our proposed method achieves a 38.37\% rank-1 matching rate, outperforming the previous best result that of SCNCD \cite{yang2014salient}, which achieved 37.8\%. The other best performing methods on the VIPeR dataset are metric learning algorithms such as kLFDA and MFA \cite{xiong2014person}. Our method performs best over ranks 1, 5, and 10, whereas kLFDA is the best at rank-20. Our experimental results suggest that even though our model suffers from a severe lack of training data, it still achieves state-of-the-art performance on the highly challenging VIPeR dataset.

Note that several of the methods considered have been combined with other descriptors to achieve better performance, such as eSDC \cite{zhao2013unsupervised} and eBiCov \cite{ma2012bicov}. To the best of our knowledge, the current best result on the VIPeR dataset is that achieved by a combination of MLF and LADF \cite{zhao2014learning}. Leveraging the considerable complementarity between the traditional framework and deep networks \cite{wei2014cnn}, we also report the results of our approach in combination with existing metric learning approaches with hand-crafted low-level features. Here, we simply sum up the scores of our method and kLFDA under the same training/testing partitions, and recompute the CMC curve. As shown in Figure \ref{subfig:cmc_viper} and Table \ref{tab:cmc_viper}, the rank-1 matching rate surges to about 53\%, far surpassing all of the state-of-the-art methods considered.

\begin{table}[!htb]
\begin{center}
\begin{tabular}{c||c|c|c|c}
  \hline
  Method & $r=1$ & $r=5$ & $r=10$ & $r=20$ \\
  \hline\hline
ELF \cite{gray2008viewpoint} & 12.00 & 41.50 & 59.50 & 74.50\\

SDALF \cite{farenzena2010person} & 19.87 & 38.89 & 49.37 & 65.73\\

CPS \cite{cheng2011custom} & 21.84 & 44.00 & 57.21 & 71.00\\

RDC \cite{zheng2013reidentification} & 15.66 & 38.42 & 53.86 & 70.09\\

aPRDC \cite{liu2012person} & 16.14 & 37.72 & 50.98 & 65.95\\

RankSVM \cite{prosser2010person} & 14.00 & 37.00 & 51.00 & 67.00\\

KISSME \cite{kostinger2012large} & 19.60 & 48.00 & 62.20 & 77.00\\

PCCA \cite{mignon2012pcca} & 19.27 & 48.89 & 64.91 & 80.28\\

rPCCA \cite{xiong2014person} & 21.96 & 54.78 & 70.97 & 85.29\\

eBiCov \cite{ma2012bicov} & 20.66 & 42.00 & 56.18 & 68.00\\

LMNN-R \cite{dikmen2011pedestrian} & 20.00 & 49.00 & 66.00 & 79.00\\

eSDC \cite{zhao2013unsupervised} & 26.31 & 46.61 & 58.86 & 72.77\\

SalMatch \cite{zhao2013person} & 30.16 & 52.31 & 65.54 & 79.15\\

MLF \cite{zhao2014learning} & 29.11 & 52.34 & 65.95 & 79.87\\

LF \cite{pedagadi2013local} & 24.18 & 52.00 & 67.12 & 82.00\\

LADF \cite{li2013learning} & 29.34 & 61.04 & 75.98 & 88.10\\

MFA \cite{xiong2014person} & 32.24 & 65.99 & 79.66 & 90.64\\

kLFDA \cite{xiong2014person} & 32.33 & 65.78 & 79.72 & \textbf{90.95}\\

SCNCD \cite{yang2014salient} & 37.80 & 68.67 & 81.01 & 90.51\\
\hline
\textbf{Ours} & \textbf{38.37} & \textbf{69.22} & \textbf{81.33} & 90.43\\
\hline\hline
MLF + LADF \cite{zhao2014learning} & 43.39 & 73.04 & 84.87 & 93.70\\
\hline
\textbf{Ours + kLFDA} & \textbf{52.85} & \textbf{81.96} & \textbf{90.51} & \textbf{95.73} \\
\hline
\end{tabular}
\end{center}
\caption{Top-ranked matching rates (\%) on the VIPeR dataset ($p=316$). The best results are highlighted in \textbf{bold}.}\label{tab:cmc_viper}
\end{table}

\begin{table}[!htb]
\begin{center}
\begin{tabular}{c||c|c|c|c}
  \hline
  Method & $r=1$ & $r=5$ & $r=10$ & $r=20$ \\
  \hline\hline
$\ell_1$-norm \cite{zhao2013person} & 10.33 & 20.64 & 26.34 & 33.52\\

$\ell_2$-norm \cite{zhao2013person} & 9.84 & 19.84 & 26.42 & 33.13\\

SDALF \cite{farenzena2010person} & 9.90 & 22.57 & 30.33 & 41.03\\

eSDC \cite{zhao2013unsupervised} & 19.67 & 32.72 & 40.29 & 50.58\\

LMNN \cite{weinberger2009distance} & 13.45 & 31.33 & 42.25 & 54.11\\

ITML \cite{davis2007information} & 15.98 & 35.22 & 45.60 & 59.81\\

Generic Metric \cite{li2012human} & 20.00 & 43.58 & 56.04 & 69.27\\

SalMatch \cite{zhao2013person} & 28.45 & 45.85 & 55.67 & 67.95\\

MLF \cite{zhao2014learning} & 34.30 & 55.06 & 64.96 & 74.94\\

kLFDA & 32.76 & 59.01 & 69.63 & 79.18\\
\hline
\textbf{Ours} & \textbf{50.41} & \textbf{75.93} & \textbf{84.07} & \textbf{91.32}\\
\hline\hline
\textbf{Ours+kLFDA} & \textbf{57.28} & \textbf{81.07} & \textbf{88.44} & \textbf{93.46} \\
\hline
\end{tabular}
\end{center}
\caption{Top-ranked matching rates (\%) on the CUHK-01 dataset ($p=486$). The best results are highlighted in \textbf{bold}.}\label{tab:cmc_cuhk}
\end{table}

\begin{table}[!htb]
\begin{center}
\begin{tabular}{c||c|c|c|c}
  \hline
  Method & $r=1$ & $r=5$ & $r=10$ & $r=20$ \\
  \hline\hline
  FW ($N=5$) \cite{martinel2015re} & 33.20 & 78.50 & 94.10 & \textbf{100.00}\\
  LF ($N=5$) \cite{pedagadi2013local} & 36.19 & 66.15 & 88.56 & 98.41\\
  kLFDA ($N=5$) & 35.20 & 75.60 & 90.96 & 99.76\\
  \hline
  \textbf{Ours} ($N=5$) & \textbf{42.24} & \textbf{82.48} & \textbf{94.72} & 99.92\\
  \hline\hline
  ICT ($N=10$) \cite{avraham2012learning} & 26.80 & 70.40 & 90.00 & 99.60\\
  FW ($N=10$) \cite{martinel2015re} & 41.90 & 86.50 & 96.70 & 100.00\\
  kLFDA ($N=10$) & 40.00 & 77.60 & 94.40 & 100.00\\
  \hline
  \textbf{Ours} ($N=10$) & \textbf{47.20} & \textbf{87.20} & \textbf{97.60} & \textbf{100.00}\\
  \hline\hline
  \textbf{Ours+kLFDA} ($N=10$) & \textbf{53.60} & \textbf{88.00} & 96.00 & \textbf{100.00}\\
  \hline
  \end{tabular}
  \end{center}
  \caption{Top-ranked matching rates (\%) on the CAVIAR4REID dataset ($p=25$). The best results are highlighted in \textbf{bold}.}\label{tab:cmc_caviar}
\end{table}

\subsubsection{Performance on the CUHK-01 dataset}

The CUHK-01 dataset \cite{li2012human} is larger in scale than the VIPeR. It contains 971 persons, each of whom has two images in each camera view. Camera \emph{A} captures the frontal or back views of pedestrians, whereas camera \emph{B} captures their side views. All images are normalized to $160\times60$. Note that we used two images of each person for training, and randomly selected only one for the test.

We compared our proposed method with several state-of-the-art approaches, such as MLF \cite{zhao2014learning} and SalMatch \cite{zhao2013person}. The CMC curves obtained using the $\ell_1$-norm and $\ell_2$-norm distances of concatenated dense features were also compared as baselines \cite{zhao2013person}. As shown in Figure \ref{subfig:cmc_cuhk} and Table \ref{tab:cmc_cuhk}, our method outdistances all state-of-the-art methods at all ranks, which again validates its effectiveness. Our method achieves a rank-1 matching rate of 50.41\%, outperforming the previous best result reported by MLF, which achieved a 34.30\% rank-1 matching rate, by a sizeable margin. The significant advantage of our proposed method is that more training data are fed to the deep network to learn a data-driven solution for the specific scenario in question.

As before, we also combined our method with kLFDA. Note that no previous work has reported the CMC curve of kLFDA on the CUHK-01 dataset, and hence we conducted this experiment with the codes offered by \cite{sugiyama2007dimensionality} and commonly used low-level features. Similar to \cite{zheng2013reidentification}, we extracted two types of low-level features: color and texture for each image. More specifically, we equally partitioned each image into $N$ horizontal stripes. For each stripe, color hisograms and Gabor texture features were extracted. Each feature channel was represented as an $\ell_1$-normalized 16-bin histogram, and all histograms were concatenated to form a single feature vector. As shown in Figure \ref{subfig:cmc_cuhk} and Table \ref{tab:cmc_cuhk}, the combination of our approach with kLFDA improves the rank-1 matching rate by about 7\%.

\subsubsection{Performance on the CAVIAR4REID dataset}

The CAVIAR4REID dataset is composed of 1,220 images of 72 pedestrians out of which 50 are viewed by two disjoint cameras. It has broad changes in resolution, the minimum and maximum size of the images is $17\times39$ and $72\times144$, respectively. It is noteworthy that both of the VIPeR and CUHK-01 are kind of regular, in the sense that pedestrians are rigidly enclosed under fixed-size bounding boxes, while CAVIAR4REID teems with significant variations in image resolution. Thus, we conducted experiment on this dataset in order to verify whether resizing would impact the proposed method.

Following \cite{martinel2015re}, we considered only 50 persons viewed by two cameras, and discarded the remaining 22 persons who appear in only one camera. In this way, the 50 persons are equally divided into training and test sets of 25 persons each. We compare our method with state-of-the-art approaches in the multi-shot modality with both $N=5$ and $N=10$. Since kLFDA performs better on both VIPeR and CUHK-01 datasets, we conducted additional experiment and also compare the CMC curves obtained by kLFDA as before. Figure \ref{subfig:cmc_caviar} and Table \ref{tab:cmc_caviar} present the results, showing that our proposed method outperforms all previous methods in both settings with $N=5$ and $N=10$, respectively. Therefore, we can conclude that our proposed method is also robust to severe variation in image resolution. In addition, as expected, the combination of our method with kLFDA boosts the rank-1 matching rate by over 6\%.

These comparisons clearly suggest that our proposed method outperforms all state-of-the-art algorithms, particularly when sufficient training data are provided. The main reason for its superior performance is that our framework is capable of jointly tackling representation learning and the learning-to-rank task rather than requiring two-step separate optimization.

\subsection{Comparision with CNN-based Person Re-identification Algorithms} \label{ssec:cmp_deepreid}
\begin{figure}[t]
  \centering
  \subfigure[]
  {\label{subfig:cmp_dml}
  \includegraphics[width=0.23\textwidth, angle=0]{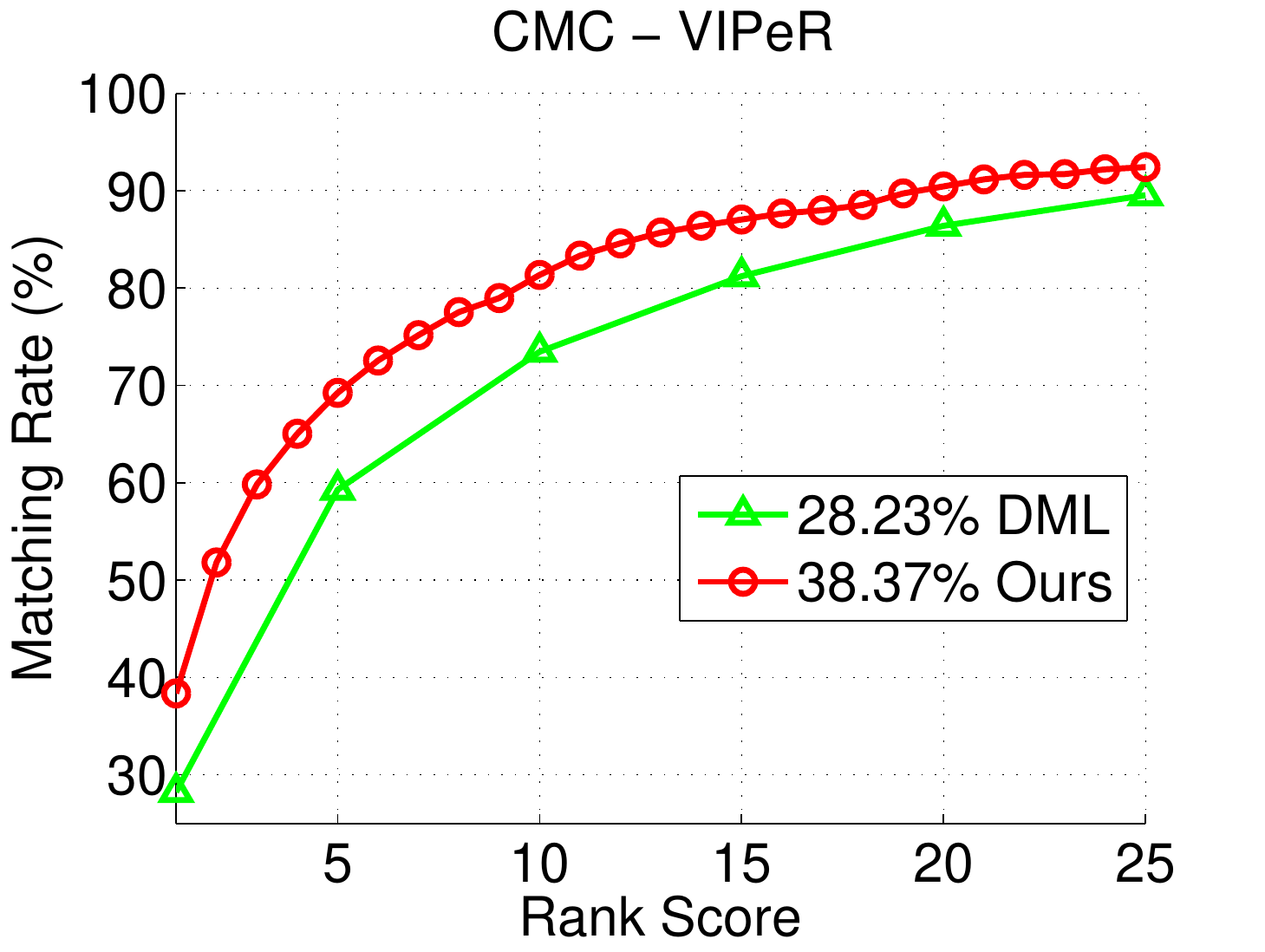}}
  \subfigure[]
  {\label{subfig:cmp_fpnn}
  \includegraphics[width=0.23\textwidth, angle=0]{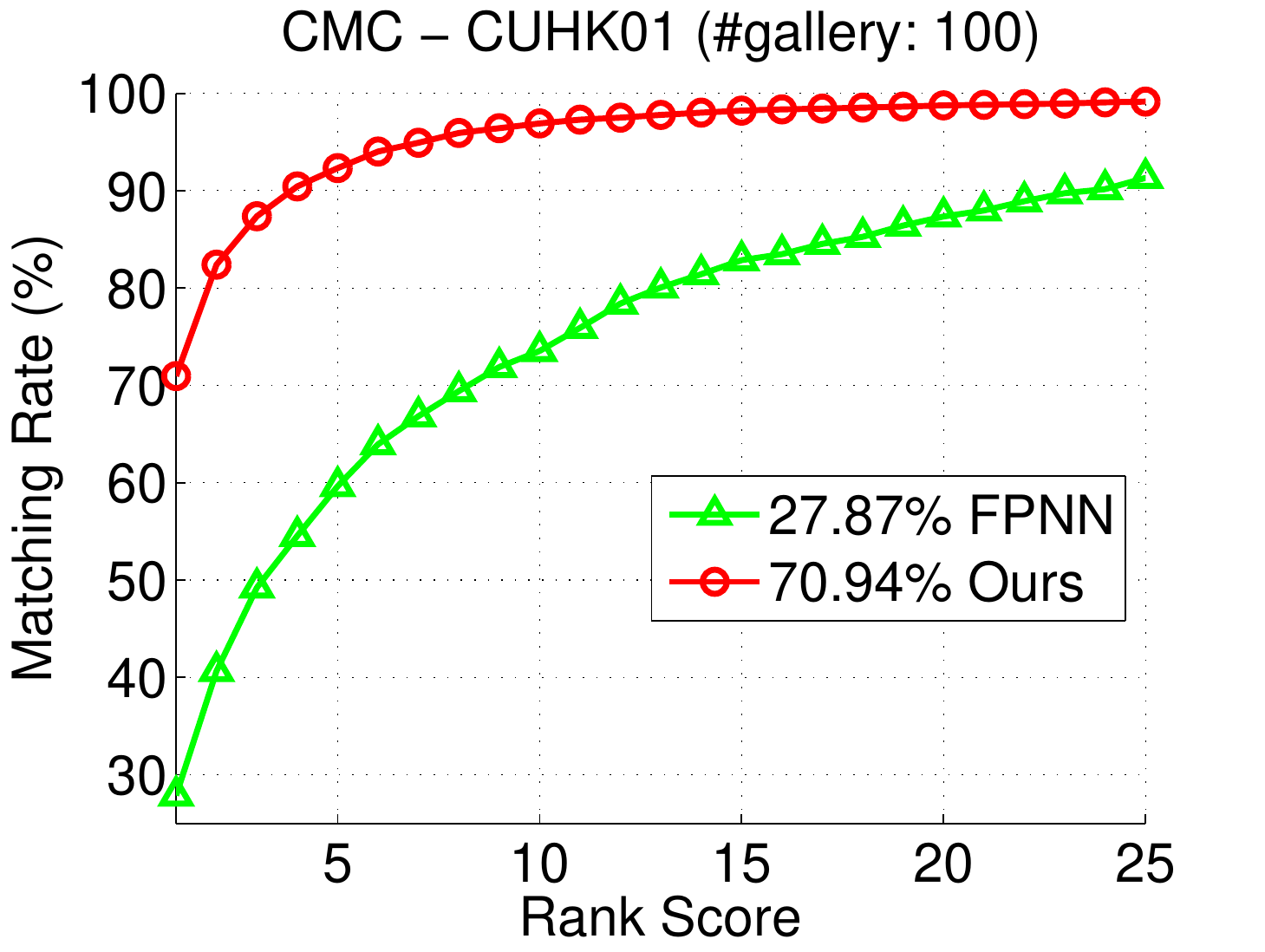}}
  \caption{Comparison with two deep learning based methods: DML \cite{yi2014deep} and FPNN \cite{li2014deepreid}. (a) Comparison with DML on VIPeR ($p=316$); (b) Comparison with FPNN on CUHK-01 ($p=100$). Note that the gallery contains only 100 subjects, which is different from that in Figure \ref{subfig:cmc_cuhk} and Table \ref{tab:cmc_cuhk}.}
  \label{fig:cmp_deepreid}
\end{figure}
In this section, we present the results of comparison between our method and two deep learning based person re-identification algorithms: DML \cite{yi2014deep} and deep FPNN \cite{li2014deepreid}. DML has been used only in experiments on the VIPeR dataset. Li \emph{et al.} \cite{li2014deepreid} concluded that existing datasets are too small to train deep networks, and thus they only conducted their experiments on both the large-scale CUHK-03 dataset and the CUHK-01 dataset. No previous CNN-based algorithm reported the results on the CAVIAR4REID. Therefore, we compare our method with DML on VIPeR and with FPNN on CUHK-01. Table \ref{tab:cmp_dml_and_fpnn}(a) and Figure \ref{subfig:cmp_dml} suggest that our model significantly surpasses DML, particularly at rank-1 (over 10\%). Note that the FPNN experiment on CUHK-01 was conducted in a different setting, with only 100 persons chosen for testing and the remaining 871 persons used for training and validation. Recall that we used 486 persons for testing and only 485 for training. In other words, we used an approximately 5-fold larger gallery than that in \cite{li2014deepreid}. Even though our setting was much more challenging than the FPNN setting, we still achieve a 50.41\% rank-1 matching rate, far surpassing that of FPNN, which was only 27.87\%. For a fairer comparison, we randomly removed 386 persons from the gallery and recomputed the average CMC curve, as shown in Figure \ref{subfig:cmp_fpnn} and Table \ref{tab:cmp_dml_and_fpnn}(b). Our proposed method achieves a 70.94\% rank-1 matching rate, a greater than 43\% improvement over FPNN.

Note that DML, FPNN, and our proposed method learn feature representation with deep CNNs, and hence their performance relies primarily on the ranking algorithm rather than the representation power of CNNs. These experimental results show that our ranking mechanism model is more suitable than others for person re-identification.

\begin{table}[!htb]
\centering
\subtable[Comparison with DML on the VIPeR dataset]{\label{subtab:cmp_dml}
\begin{tabular}{c||c|c|c|c|c}
  \hline
  Method & $r=1$ & $r=5$ & $r=10$ & $r=20$ & $r=30$\\
  \hline\hline
  DML \cite{yi2014deep} & 28.23 & 59.27 & 73.45 & 86.39 & 92.28\\
  \hline
  \textbf{Ours} & \textbf{38.37} & \textbf{69.22} & \textbf{81.33} & \textbf{90.43} & \textbf{94.15}\\
\hline
\end{tabular}
}
\subtable[Comparison with FPNN on the CUHK-01 dataset]{\label{subtab:cmp_fpnn}
\begin{tabular}{c||c|c|c|c|c}
  \hline
  Method & $r=1$ & $r=5$ & $r=10$ & $r=20$ & $r=30$\\
  \hline\hline
  FPNN \cite{li2014deepreid} & 27.87 & 59.64 & 73.53 & 87.34 & 93.92\\
  \hline
  \textbf{Ours} & \textbf{70.94} & \textbf{92.30} & \textbf{96.90} & \textbf{98.74} & \textbf{99.34}\\
\hline
\end{tabular}
}
\caption{Comparison with two deep learning based person re-identification algorithms: DML \cite{yi2014deep} and FPNN \cite{li2014deepreid}. The best results are highlighted in \textbf{bold}.}
\label{tab:cmp_dml_and_fpnn}
\end{table}

\subsection{Evaluation of Open-World Scenarios}

\begin{figure*}
  \centering
  \subfigure[]
  {\label{subfig:ow_viper_6}
  \includegraphics[width=0.23\textwidth, angle=0]{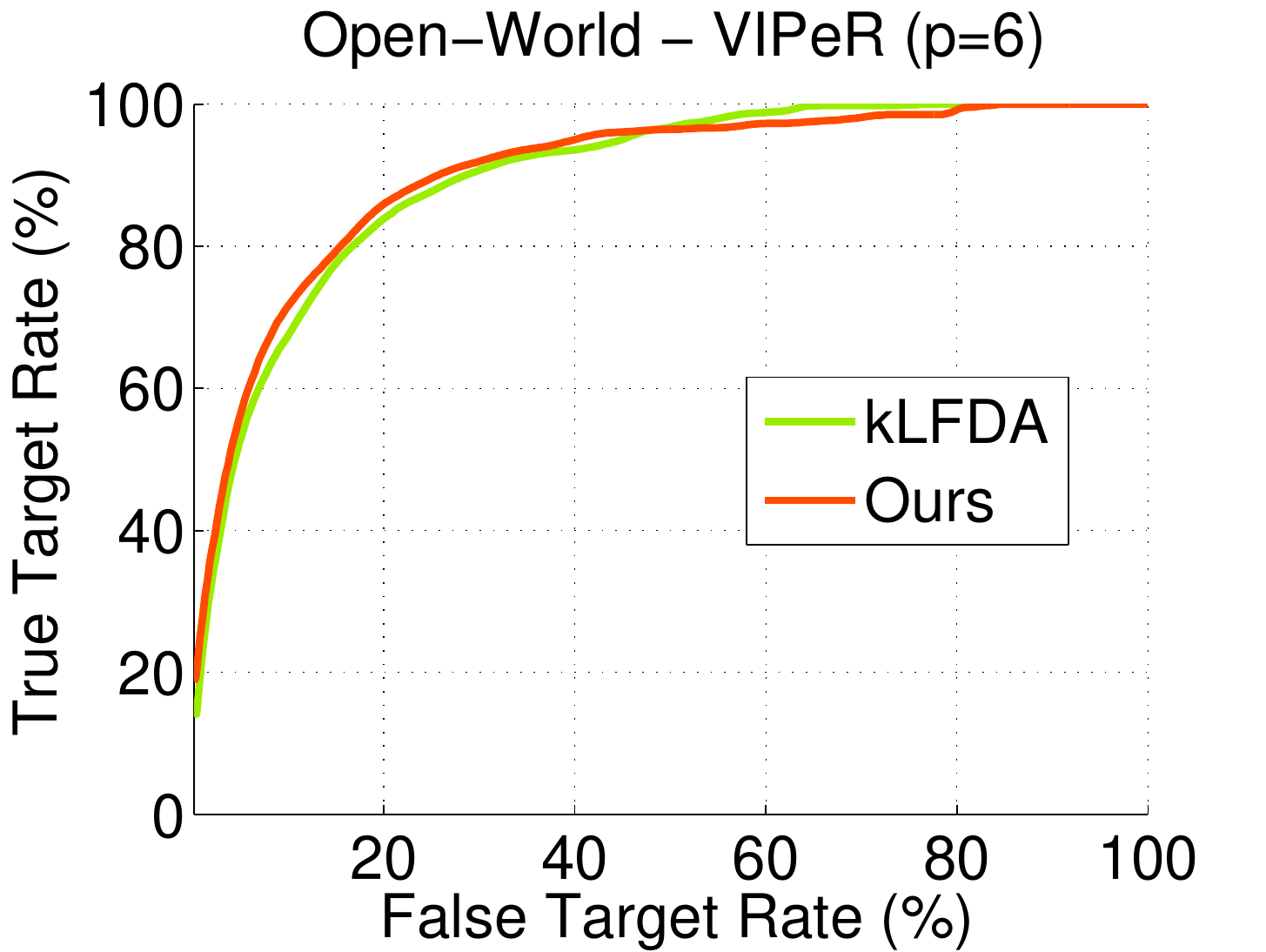}}
  \subfigure[]
  {\label{subfig:ow_viper_10}
  \includegraphics[width=0.23\textwidth, angle=0]{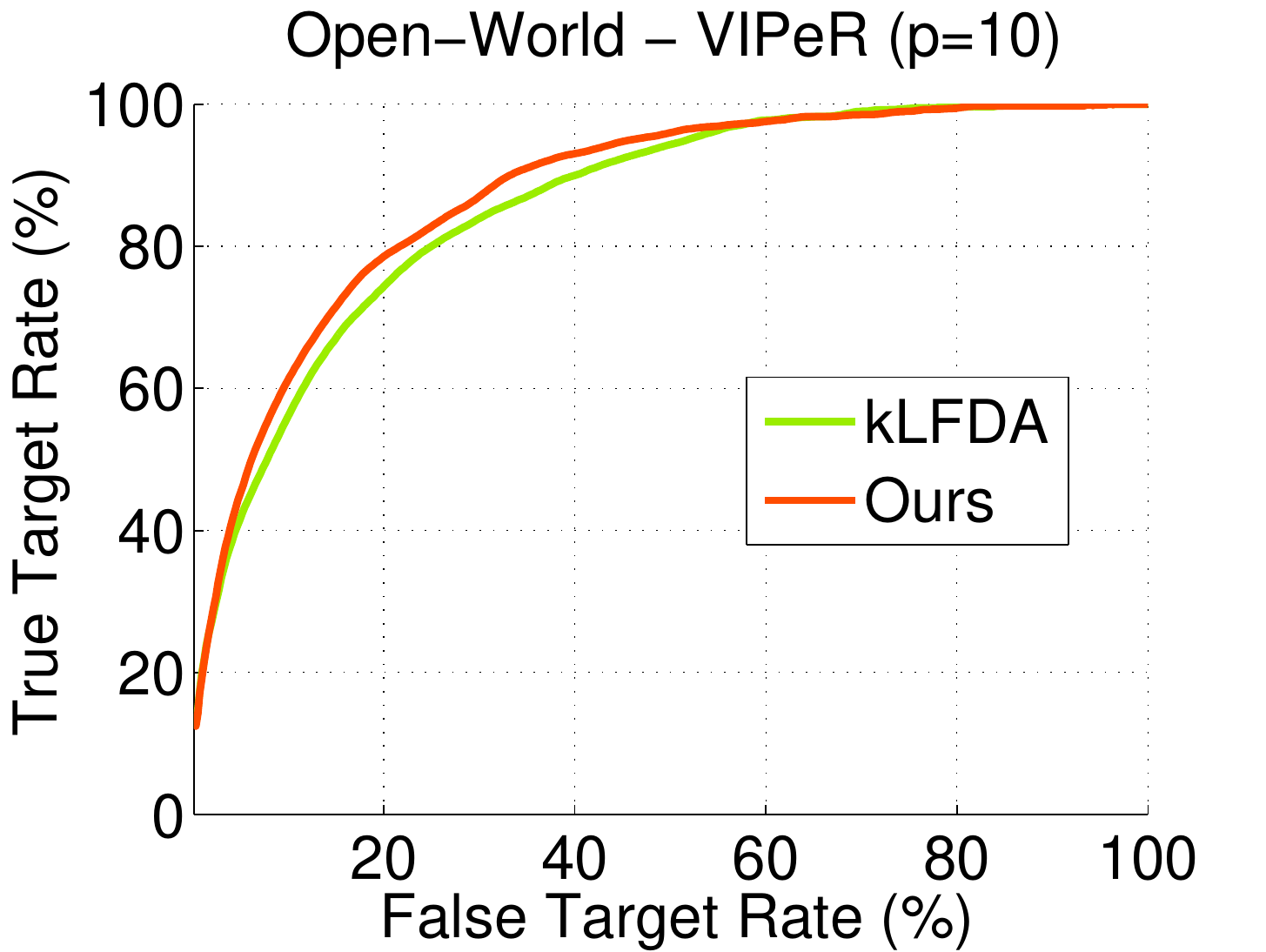}}
  \subfigure[]
  {\label{subfig:ow_cuhk01_6}
  \includegraphics[width=0.23\textwidth, angle=0]{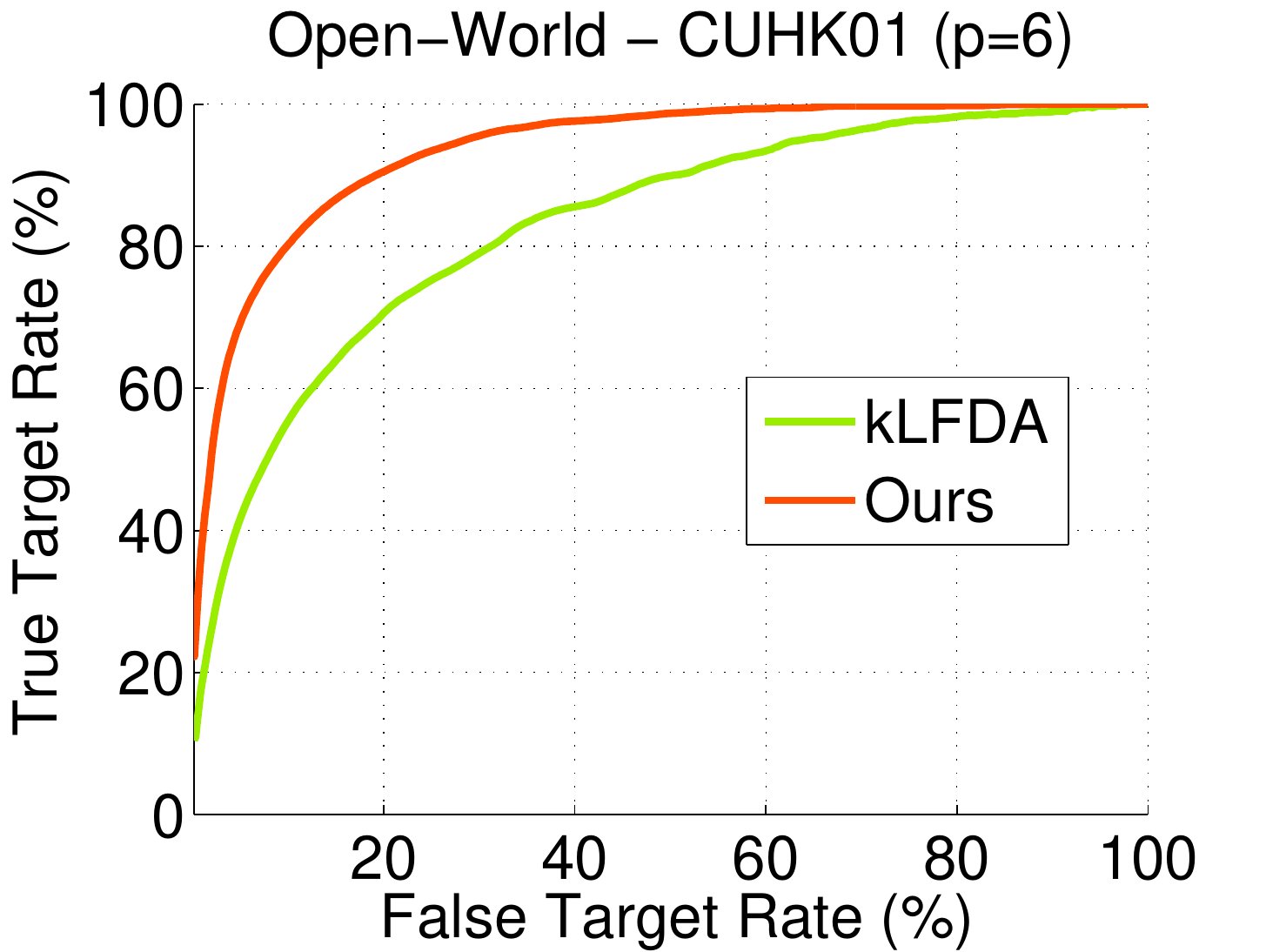}}
  \subfigure[]
  {\label{subfig:ow_cuhk01_10}
  \includegraphics[width=0.23\textwidth, angle=0]{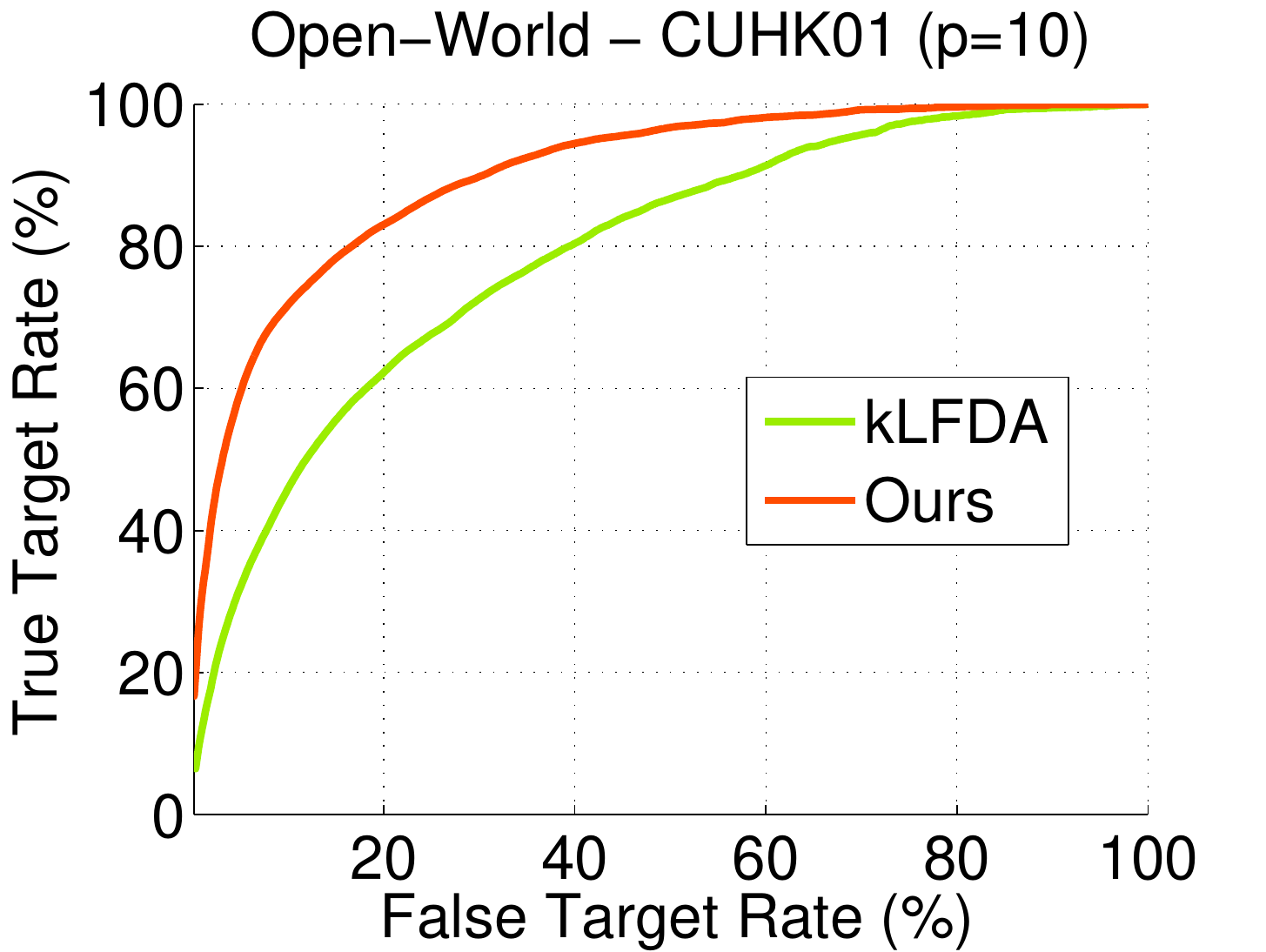}}
  \caption{Open-world set-based verification performance comparisons with kLFDA using TTR-FTR curves. Best viewed in color.}
  \label{fig:cmp_open_world}
\end{figure*}

\begin{table*}[!htb]
\begin{center}
\begin{tabular}{c||c|c|c|c|c|c||c|c|c|c|c|c}
  \hline
  Dataset & \multicolumn{6}{c||}{VIPeR ($p=6$)} & \multicolumn{6}{c}{VIPeR ($p=10$)}\\
  \hline
  FTR & $0.1\%$ & $1\%$ & $5\%$ & $10\%$ & $20\%$ & $50\%$ & $0.1\%$ & $1\%$ & $5\%$ & $10\%$ & $20\%$ & $50\%$\\
  \hline
  kLFDA & 14.58 & 23.96 & 52.71 & 67.71 & 83.96 & \textbf{96.67} & 12.88 & 18.88 & 42.63 & 56.38 & 74.50 & 94.38 \\
  \textbf{Ours} & \textbf{18.75} & \textbf{29.79} & \textbf{58.33} & \textbf{71.46} & \textbf{86.25} & 96.46 & \textbf{13.88} & \textbf{19.38} & \textbf{46.25} & \textbf{61.00} & \textbf{78.50} & \textbf{95.88} \\
  \hline\hline
  Dataset & \multicolumn{6}{c||}{CUHK-01 ($p=6$)} & \multicolumn{6}{c}{CUHK-01 ($p=10$)}\\
  \hline
  FTR & $0.1\%$ & $1\%$ & $5\%$ & $10\%$ & $20\%$ & $50\%$ & $0.1\%$ & $1\%$ & $5\%$ & $10\%$ & $20\%$ & $50\%$\\
  \hline
  kLFDA & 10.53 & 19.13 & 42.40 & 55.60 & 70.87 & 89.93 & 6.80 & 12.52 & 32.72 & 46.04 & 62.28 & 86.68 \\
  \textbf{Ours} & \textbf{22.87} & \textbf{40.13} & \textbf{69.53} & \textbf{80.07} & \textbf{90.60} & \textbf{98.73} & \textbf{14.92} & \textbf{32.84} & \textbf{59.48} & \textbf{71.96} & \textbf{83.04} & \textbf{96.72} \\
  \hline

  \end{tabular}
  \end{center}
  \caption{Open-World set-based verification results: True Target Rate (TTR) in \% against False Target Rate (FTR). The best results are highlighted in \textbf{bold}.}\label{tab:open_world}
\end{table*}

The CMC-based evaluation protocol used in Section \ref{ssec:cmp} and \ref{ssec:cmp_deepreid} assumes that the gallery and probe sets contain exactly the same individuals. Thus, this can be regarded as the \emph{close-world} re-identification task. Here we also consider the person re-identification problem in the context of open-world scenarios \cite{zhengtowards}. The main difficulty of open-world setting lies in the fact that the probe image may not belong to anyone on the gallery/target set. Under this setting, a watch-list of a handful of known people is provided as the target set. Additionally, there are a large amount of non-target imposters captured along with the target set. Given a probe image, we need to determine whether it is on the watch-list or not.

Following the open-world evaluation metrics defined in \cite{zhengtowards}, we exploited the True Target Rate (TTR) and False Target Rate (FTR) as,
\begin{equation}
TTR = \frac{\#\mathcal{TTQ}}{\#\mathcal{TQ}},
\end{equation}
\begin{equation}
FTR = \frac{\#\mathcal{FNTQ}}{\#\mathcal{NTQ}},
\end{equation}
where $\mathcal{TQ}$ denotes the query target images from target people; $\mathcal{NTQ}$ indicates the query non-target images from non-target people; $\mathcal{TTQ}$ denotes the query target images that are verified as one of the target people; and $\mathcal{FNTQ}$ means the query non-target images that are mistakenly verified as one of the target people. In the experiment, we randomly selected $p$ subjects from the gallery set as the target set, and removed the remaining images from the gallery. In this way, most probe images cannot find their true matches in the target set, which contains several pedestrians that we are interested in. As in the close-world setting, we computed the similarity score for each probe image and each target image. A value $s$ is used to threshold these scores and therefore a curve depicting the TTR value against the FTR value is reported by changing the value $s$. We also reported the TTR value when the FTR value is fixed. We repeated this experiment 100 times and reported the average verification performance on the VIPeR and CUHK-01 datasets, respectively. The comparison of our method and kLFDA is given. Figure \ref{fig:cmp_open_world} shows the TTR-FTR curves with $p=6$ and $p=10$ (set as that in \cite{zhengtowards}). More detailed comparison of TTR values with FTR fixed is presented in Table \ref{tab:open_world}. It demonstrates that, our proposed method also surpasses kLFDA under the open-world scenarios, particularly when sufficient training samples are given.

\subsection{Comparison of Performance across Datasets}
In this section, we compare the performance in the cross dataset setting which better coincides with practical applications.
In the case of large-scale camera networks, it is impossible to collect enough labeled data to learn a specific discriminative model for each camera pair. Therefore, transferring the learnt model to the current scenario without significant degradation in performance is a more practical approach. The Domain Transfer Ranked Support Vector Machine (DTRSVM) \cite{ma2013domain} was proposed to address this problem. However, the DTRSVM still has to re-train the metric with source domain data and easily collected negative samples from target domain in a multi-task learning framework.

\begin{figure}
\centering
\includegraphics[width=0.32\textwidth, angle=0]{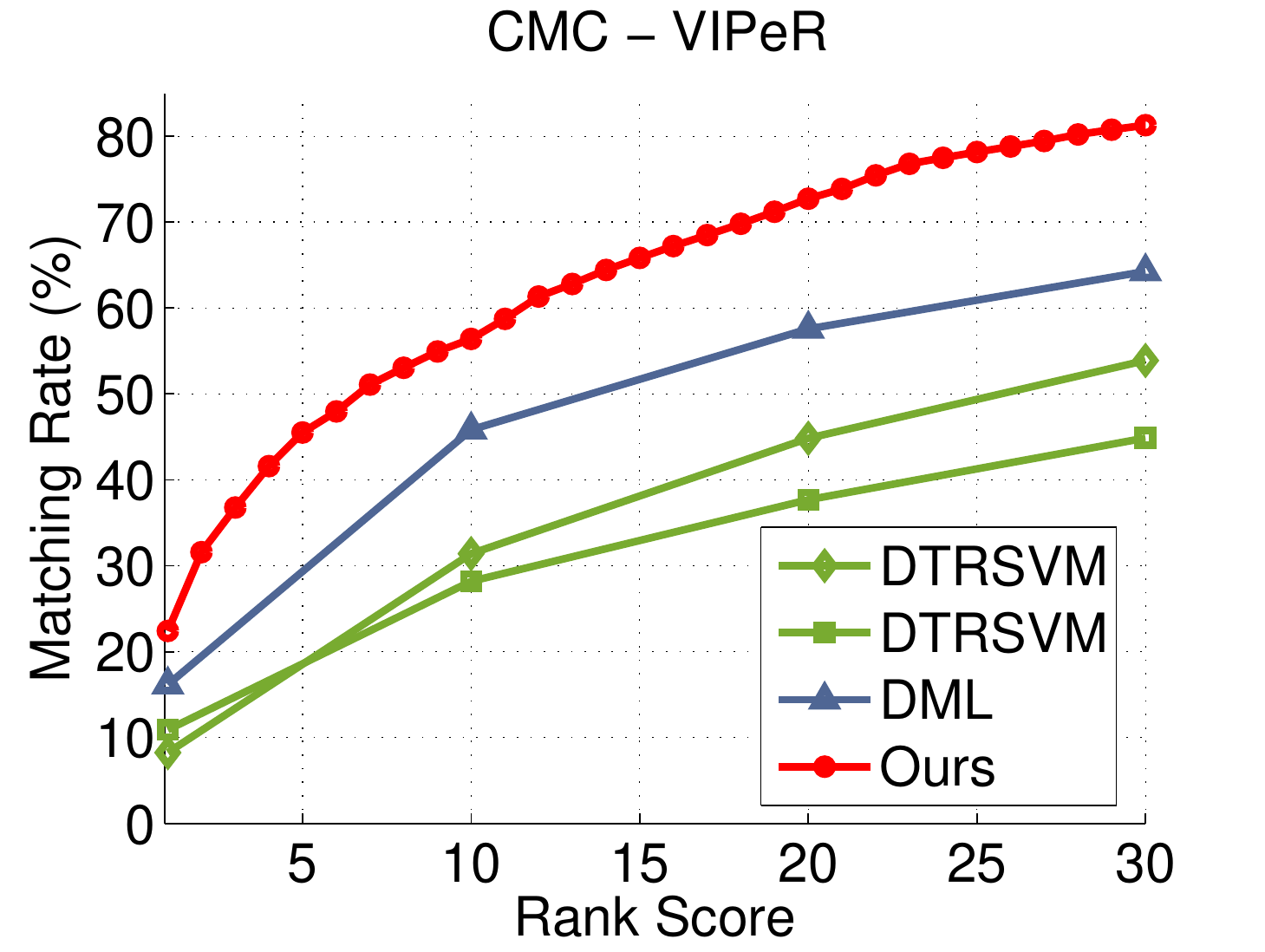}
\caption{Comparison of cross-dataset performance. Note that, different from DTRSVM, the DML and our method do not use any sample from VIPeR. Best viewed in color.}
\label{fig:cross_cmc}
\end{figure}

It is known that, deep CNNs have a strong generalization ability across datasets. Here, we compare the cross-dataset performance of our method with that of the DML \cite{yi2014deep} and the DTRSVM. Following \cite{yi2014deep}, we directly tested their performance on the VIPeR dataset with the pre-trained model learnt from CUHK-02 without fine-tuning. Note that the DTRSVM needs negative samples from VIPeR when transferring the metric, whereas our method and the DML did not use any sample on the VIPeR dataset to fine-tune the network parameters. Figure \ref{fig:cross_cmc} and Table \ref{tab:cross} summarize the CMC curves and detailed numerical matching rates, respectively. The experimental results show that: 1)~deep CNNs exhibit good generalization ability for cross-dataset re-identification tasks, even though no sample from the target domain is used to fine-tune the networks; 2)~using the same training data, our method also outperforms DML in generalization ability significantly, which again demonstrating the superiority of our proposed method.

\begin{table*}[!htb]
\begin{center}
\begin{tabular}{c||c|c|c|c|c|c}
  \hline
  Method & Training data & need to re-train & $r=1$ & $r=10$ & $r=20$ & $r=30$\\
  \hline\hline
  DTRSVM \cite{ma2013domain} & i-LIDS + VIPeR & $\surd$ & 8.26 & 31.39 & 44.83 & 53.88\\
  DTRSVM \cite{ma2013domain} & PRID + VIPeR & $\surd$ & 10.90 & 28.20 & 37.69 & 44.87\\
  DML \cite{yi2014deep} & CUHK & $\times$ & 16.17 & 45.82 & 57.56 & 64.24\\
  \hline
  \textbf{Ours} & CUHK & $\times$ & \textbf{22.41} & \textbf{56.39} & \textbf{72.72} & \textbf{81.27}\\
\hline
\end{tabular}
\end{center}
\caption{Cross-dataset experiment: Top-ranked matching rates (\%) on the VIPeR dataset. The best results are highlighted in \textbf{bold}.}\label{tab:cross}
\end{table*}

\subsection{Evaluations and Analysis}
We now analyze a fair self-evaluation of the proposed person re-identification algorithm. Fair self-evaluation is defined as evaluation of both the final output and each component of the algorithm to assess the actual contributions of various components. We are here inspired by \cite{ xiong2014person }. Unlike many works simply comparing the final CMC curves in different experimental settings, \cite{ xiong2014person } employed uniform and consistent representation in all comparisons to fairly evaluate different metric learning methods, i.e., the important components of re-identification algorithms. It is important to follow this principle, because representation and the ranking mechanism can both affect the final performance. We cannot determine the actual contribution of different ranking mechanisms, only according to the final CMC curves under different representations. Overall, the output result of a re-identification approach is determined by several key factors, mainly including pre-processing, image segmentation, pedestrian representation, ranking mechanism, etc.

In a more general setting than that in \cite{ xiong2014person }, a fair self-evaluation for a re-identification algorithm should verify that the effectiveness of the proposed algorithm stems primarily from the components that are claimed to be effective, rather than comparing only the final output or a specific component in different settings. Our aim in conducting a fair self-evaluation is to assess each component of our proposed method to prove the positive roles that all components played in our re-identification framework. Following our analysis of the extensive comparison experiments in Section \ref{ssec:cmp} and \ref{ssec:cmp_deepreid}, which demonstrate the superiority of our approach as a whole, we here evaluate and analyze each component of our deep ranking framework in detail.

\subsubsection{Contribution of joint representation learning}\label{sssec:joint}
Many previous works, including shallow and deep learning algorithms, share a similar framework: they extract features for two images separately, and then use the Euclidean or cosine distance as the metric. The only difference between them is whether the features are hand-crafted or learned by CNNs. Our approach learns joint representation for two pedestrian images and directly predicts their similarity from raw pixels, which is similar to FPNN in terms of joint representation learning, but differs from it in learning to rank (further explained below).
Our approach is motivated by human assessment: when a person assesses whether two images belong to the same pedestrian, he or she puts the two images together and compares clothes and accessories of those depicted. Actually, the discriminative information comes from different parts of the images. For instance, given three pedestrian images, \emph{a}, \emph{b}, and \emph{c}, that look quite similar, we are able to distinguish \emph{a} from \emph{b} by the subjects' knapsacks, which have different colors; and \emph{a} from \emph{c} by their different shoes. Given probe \emph{a}, the discriminative region comes from the knapsacks compared with \emph{b}, and from the shoes compared with \emph{c}. However, valuable information is often hidden or ignored when features are extracted independently. We argue that a decision is made jointly from two images rather than from two separately generated items of discriminative information. Therefore, we propose to predict the similarity of two images via joint representation learning with a learned deep network.
\begin{figure*}
  \centering
  \includegraphics[width=0.95\textwidth, angle=0]{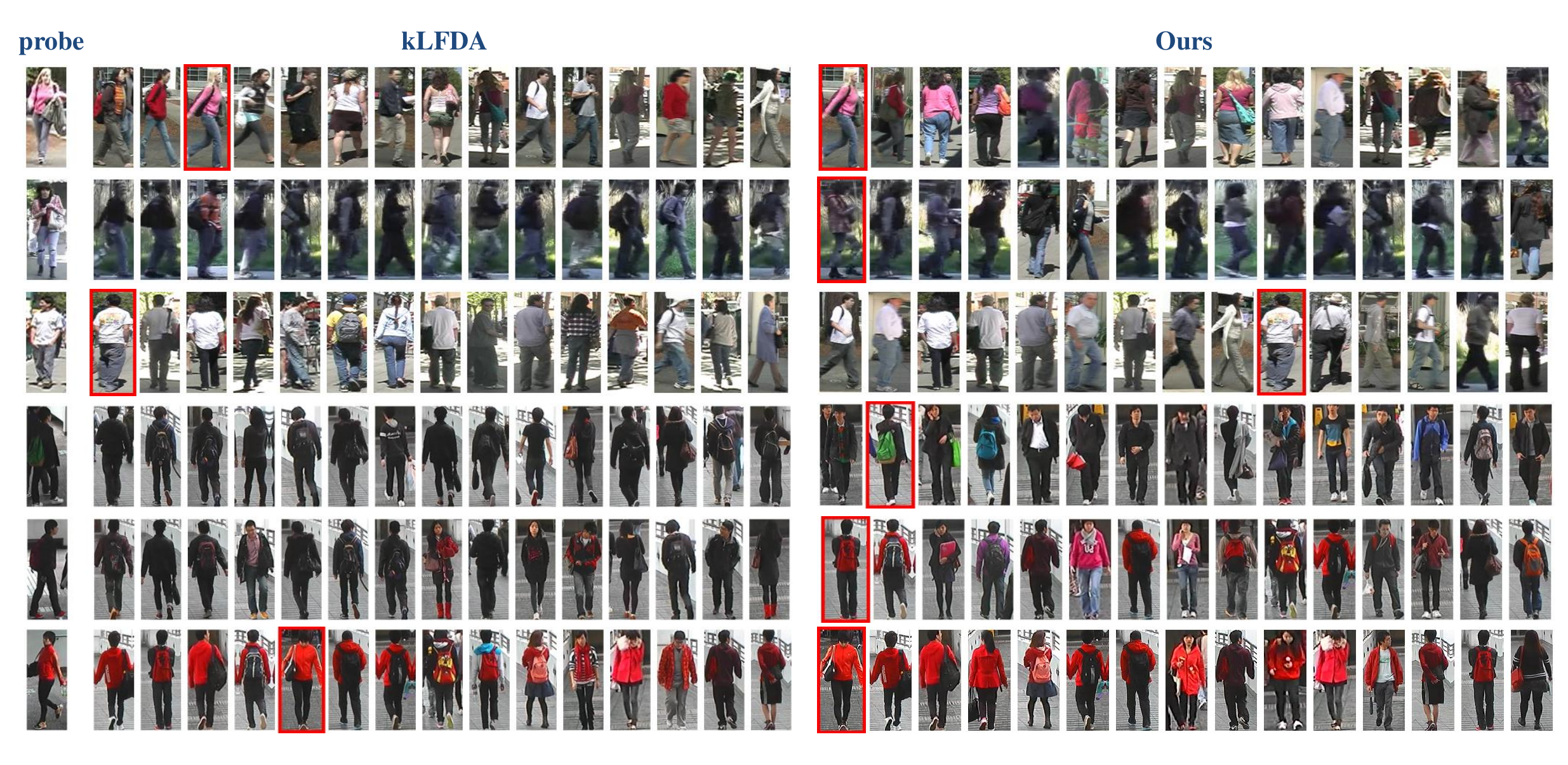}
  \caption{Comparison of the ranking examples of kLFDA and our method. In each row, the left-most image is the probe, and the others are the top 15 matched gallery images of kLFDA and our method. The correct match of the probe is highlighted with a red bounding box. Best viewed in color.}
  \label{fig:ranking_sample}
\end{figure*}

As kLFDA achieves the best performance among metric learning algorithms with hand-crafted features for re-identification at present, we compare the ranking examples of our method and those of kLFDA, as shown in Figure \ref{fig:ranking_sample}. The images in the first three rows are taken from the VIPeR dataset, and those in the last three rows come from the CUHK-01 dataset. This comparison also reveals multiple valuable facts about joint representation learning, as follows. (1)~Low-level features can easily produce counterintuitive results, whereas joint representation learning captures semantic colors very well. In the first row, given the probe dressing in pink, our method correctly places the true match at rank-1 and also places other persons wearing pink clothes in the top ranks. However, kLFDA mistakenly positions two persons who wear different colors of clothes above the true match. (2)~Joint representation learning exhibits superior performance particularly when illumination varies greatly (e.g., row 2). In rows 4 and 5, joint representation learning is able to capture discriminative information from the green or red bag in the presence of both illumination and pose variations, whereas low-level features fail. (3)~Low-level features sometimes outperform joint representation learning provided that the degree of cross-view variation is small. For instance, kLFDA performs well in row 3, because the probe shares similar pose and discriminative background to the corresponding gallery image, which just right provide a useful context cue for ranking. (4)~Row 6 is extremely challenging because several candidates look nearly the same. Our method still correctly matches the probe at rank-1, suggesting that joint representation learning is capable of mining subtle discriminative information for re-identification.

\subsubsection{Ranking versus direct binary classification}
\begin{figure}
\centering
\includegraphics[width=0.32\textwidth, angle=0]{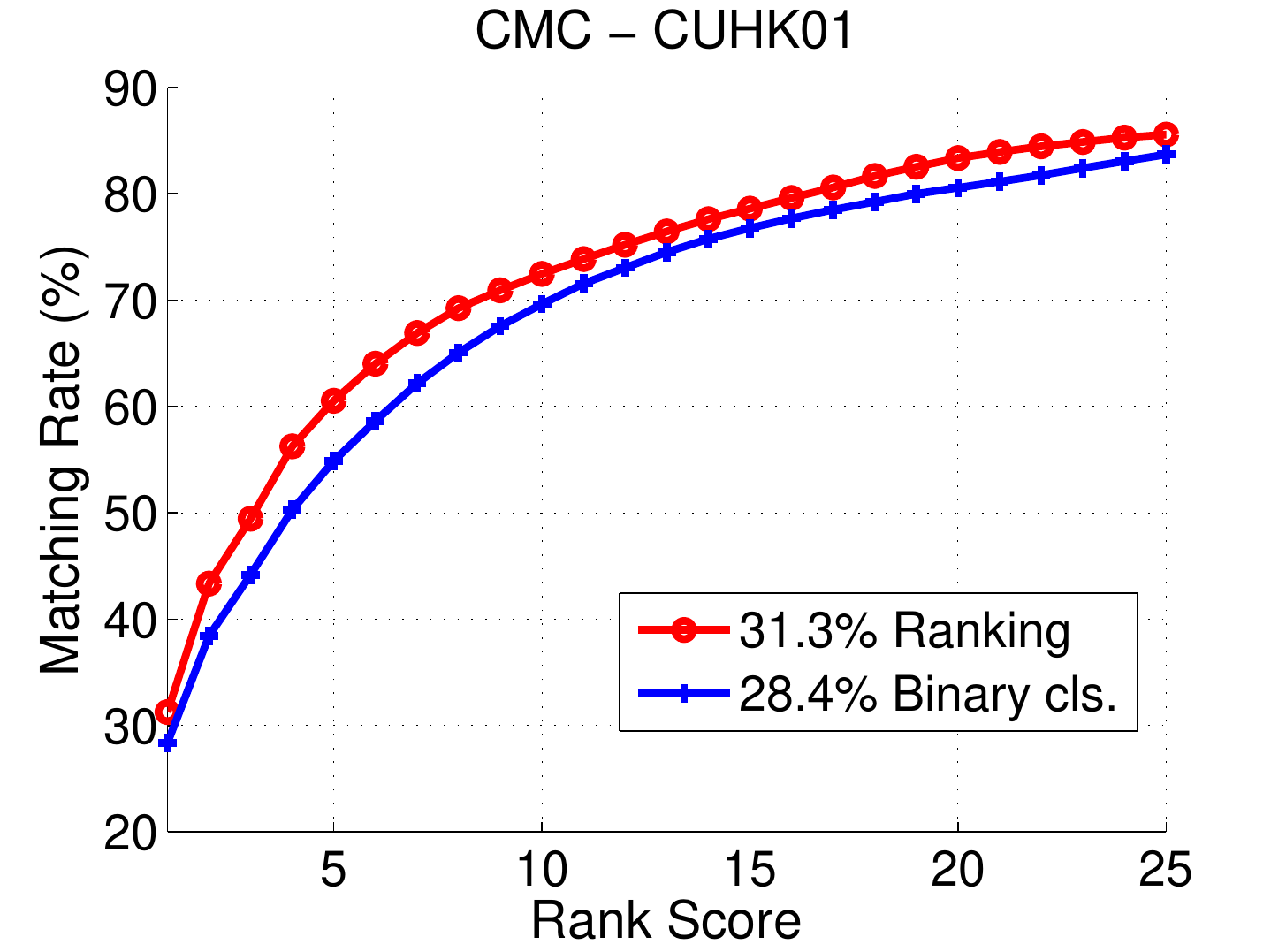}
\caption{Comparison of ranking and direct binary classification.}
\label{fig:ranking_vs_cls}
\end{figure}

We also performed experiments to evaluate the contribution of our learning-to-rank algorithm. The results reveal the strength of our ranking mechanism relative to other CNN-based methods. We first removed our proposed ranking model, and then employed a softmax layer to replace the last fully connected layer in Figure \ref{fig:net_arch}, with the other layers left unchanged. In this way, the deep network was used to assess whether two input images belonged to the same person. In other words, we performed direct binary classification instead of learning to rank for the person re-identification task like FPNN \cite{li2014deepreid}. The experiments were conducted on the CUHK-01 dataset with a positive-negative ratio of 1:1. No pre-training was used. The CMC curves in Figure \ref{fig:ranking_vs_cls} show that learning to rank consistently surpasses direct binary classification, thereby demonstrating that the good performance of our method stems from both deep representation learning and the ranking algorithm because of the intrinsic difference between image classification and ranking tasks. A good network for image classification is not optimal for ranking \cite{wang2014learning}. Our learning-to-rank algorithm is based on a relative similarity comparison rather than an absolute decision for each pair, which better accords with the spirit of person re-identification.

\subsubsection{Contribution of pre-training}
\begin{figure*}
\centering
  \subfigure[]
  {\label{subfig:cmc_pt}
  \includegraphics[width=0.32\textwidth, angle=0]{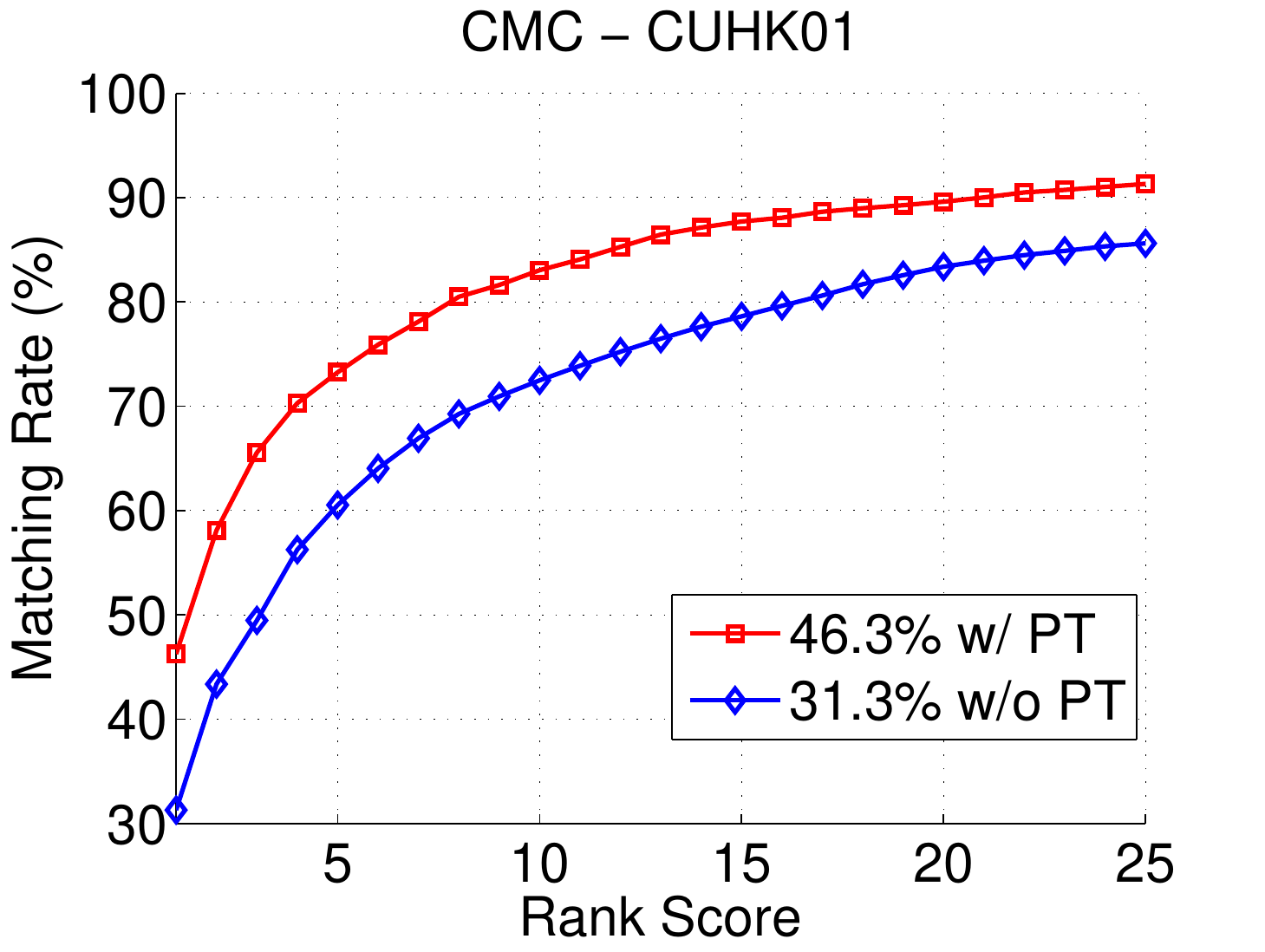}}
  \subfigure[]
  {\label{subfig:loss_wpt}
  \includegraphics[width=0.32\textwidth, angle=0]{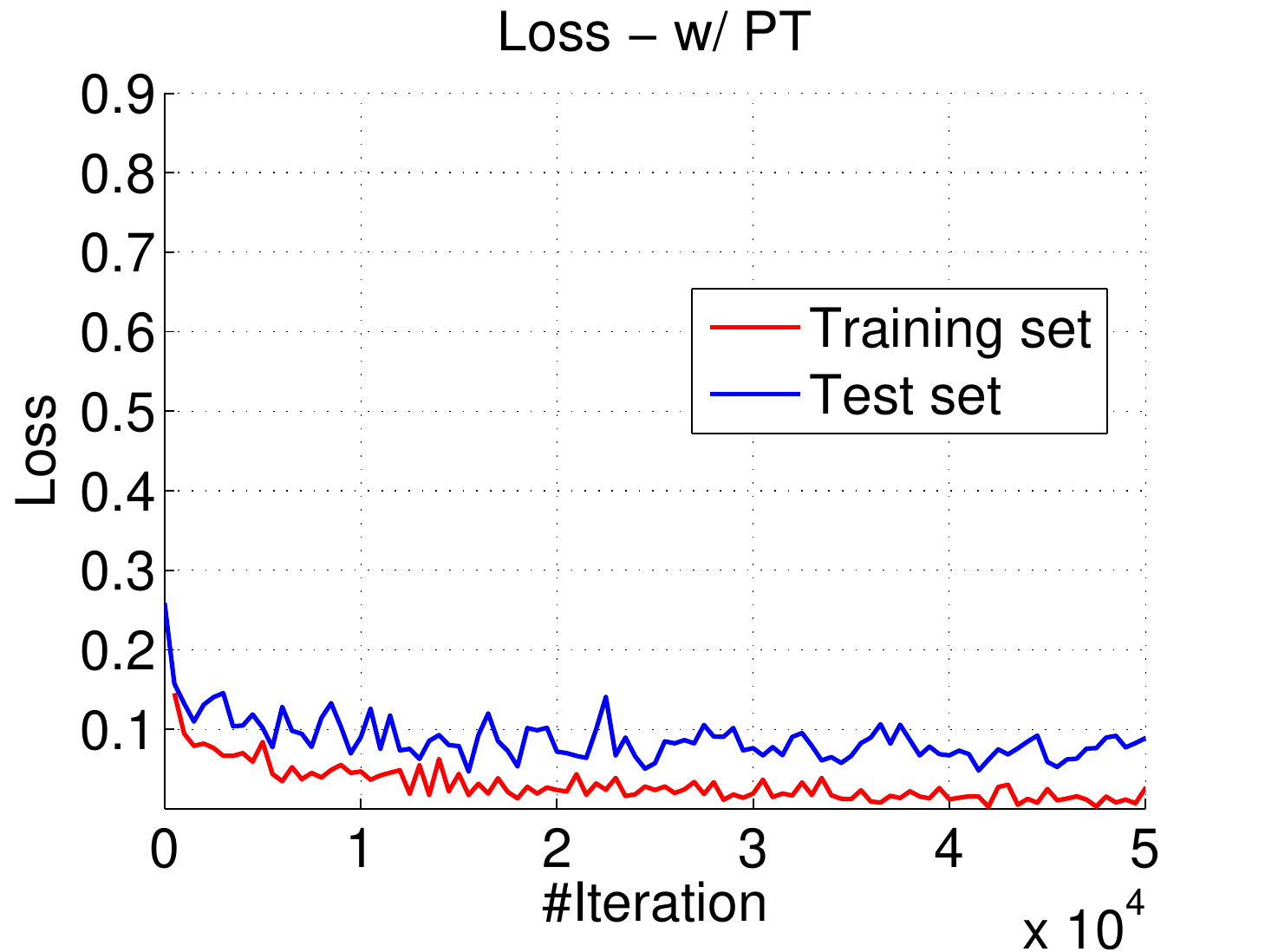}}
  \subfigure[]
  {\label{subfig:loss_wopt}
  \includegraphics[width=0.32\textwidth, angle=0]{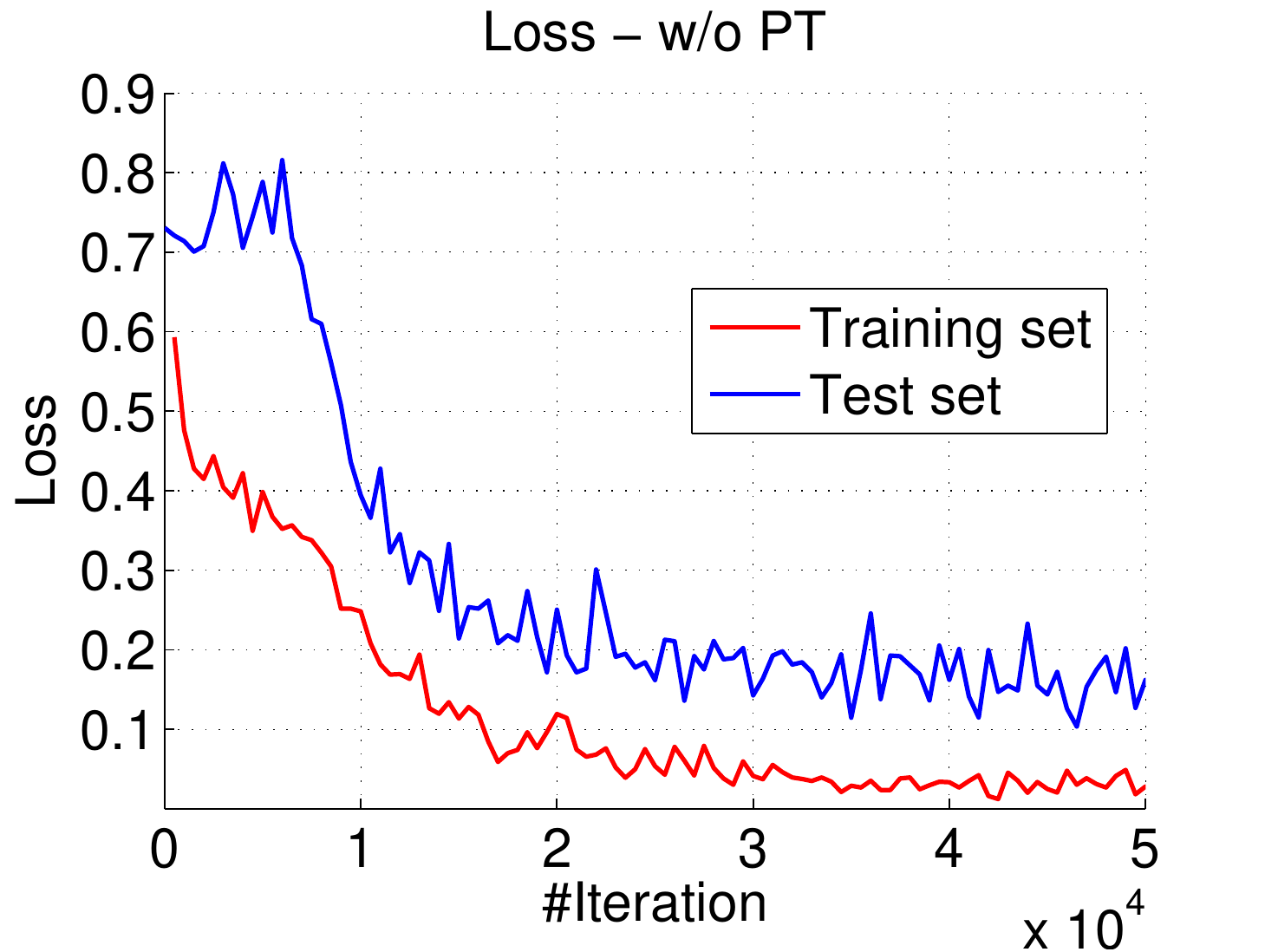}}
\caption{Analysis of the contribution of pre-training (abbreviated as PT in the figure). These experiments were conducted on the CUHK-01 dataset with a 1:1 positive-negative ratio with or without pre-training. (a) shows the CMC curves, and (b) and (c) show the loss of the training and test sets with and without pre-training, respectively. Best viewed in color.}
\label{fig:pretraining}
\end{figure*}

We evaluated the contribution of pre-training by comparing the performance with and without it. Even though it is difficult to quantify the amount of data needed with reference to the network¡¯s depth, deep networks have shown ravenous appetite for training data. As an example, the verification rate improves about 10\% for the same deep network, when the number of training images increases from 2.6 million to 26 million \cite{schroff2015facenet}. Here, we performed experiments on the CUHK-01 dataset with a positive-negative ratio of 1:1. Figure \ref{fig:pretraining} shows the CMC curves and the loss of the training and test sets. As depicted in Figure \ref{subfig:cmc_pt}, pre-training improves performance by 15\% at rank-1, and consistently boosts it by about 10\% at all ranks. Our deep network needs more training data to learn the parameters, and pre-training is thus able to make use of large-scale outside data. In addition, Figures \ref{subfig:loss_wpt} and \ref{subfig:loss_wopt} clearly show that the deep network is given better initialization and converges much faster through pre-training. The implication is that the re-identification performance of our proposed method would be further improved through a pre-trained model learned with larger-scale labeled data.

It is noteworthy that pre-training is not an indispensable component to our proposed framework. As discussed in previous section, pre-training allows us to enlarge the depth of our neural network, helping to learn more robust and discriminative representation. In this way, the size of network is no longer limited by the labeled training data collected from the target scenario. As an example, we have to design a small-sized network for the VIPeR dataset without pre-training. Fortunately, leveraging the generalization ability of CNNs, we can borrow outside data to pre-train a deeper network, and then fine-tune its parameters for the current target scenario for better re-identification performance.

\begin{figure}[!h]
\centering
\includegraphics[width=0.32\textwidth, angle=0]{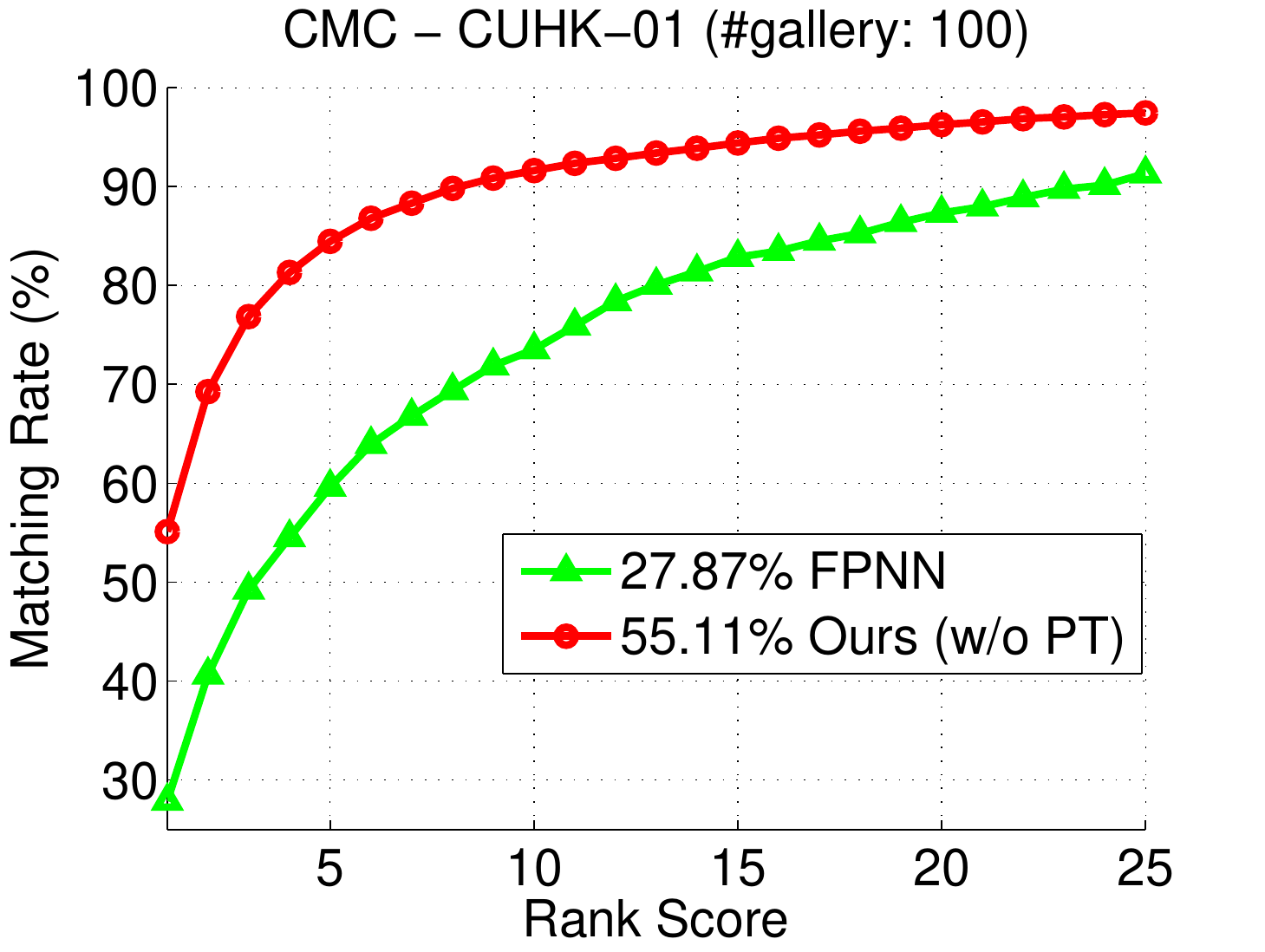}
\caption{Comparison with FPNN on the CUHK-01 dataset ($p=100$). Note that no pre-training is used here.}
\label{fig:cmp_fpnn_nopre}
\end{figure}

\begin{figure}
\centering
\subfigure[]
{\label{subfig:viper_aug}
\includegraphics[width=0.23\textwidth, angle=0]{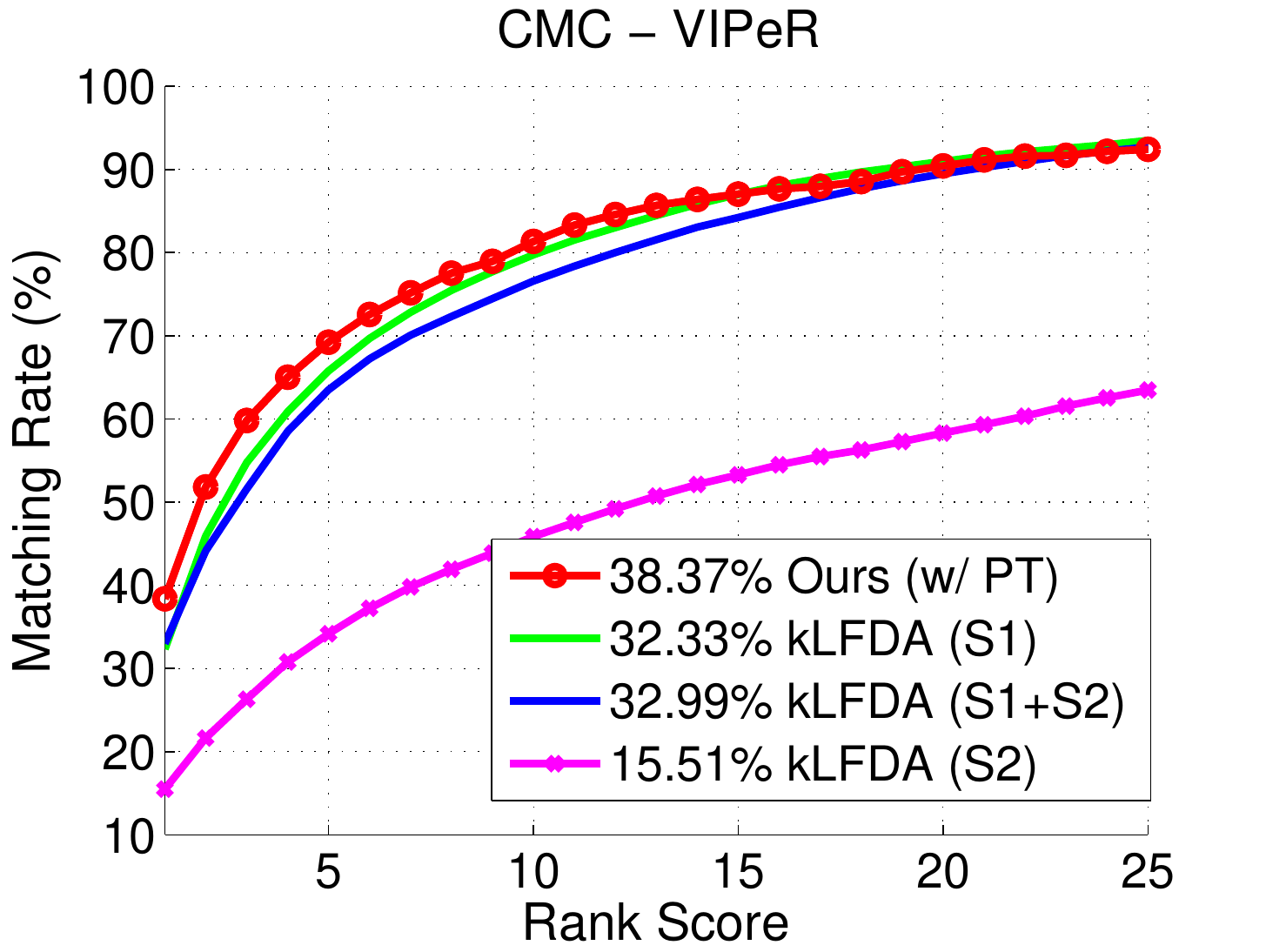}}
\subfigure[]
{\label{subfig:cuhk01_aug}
\includegraphics[width=0.23\textwidth, angle=0]{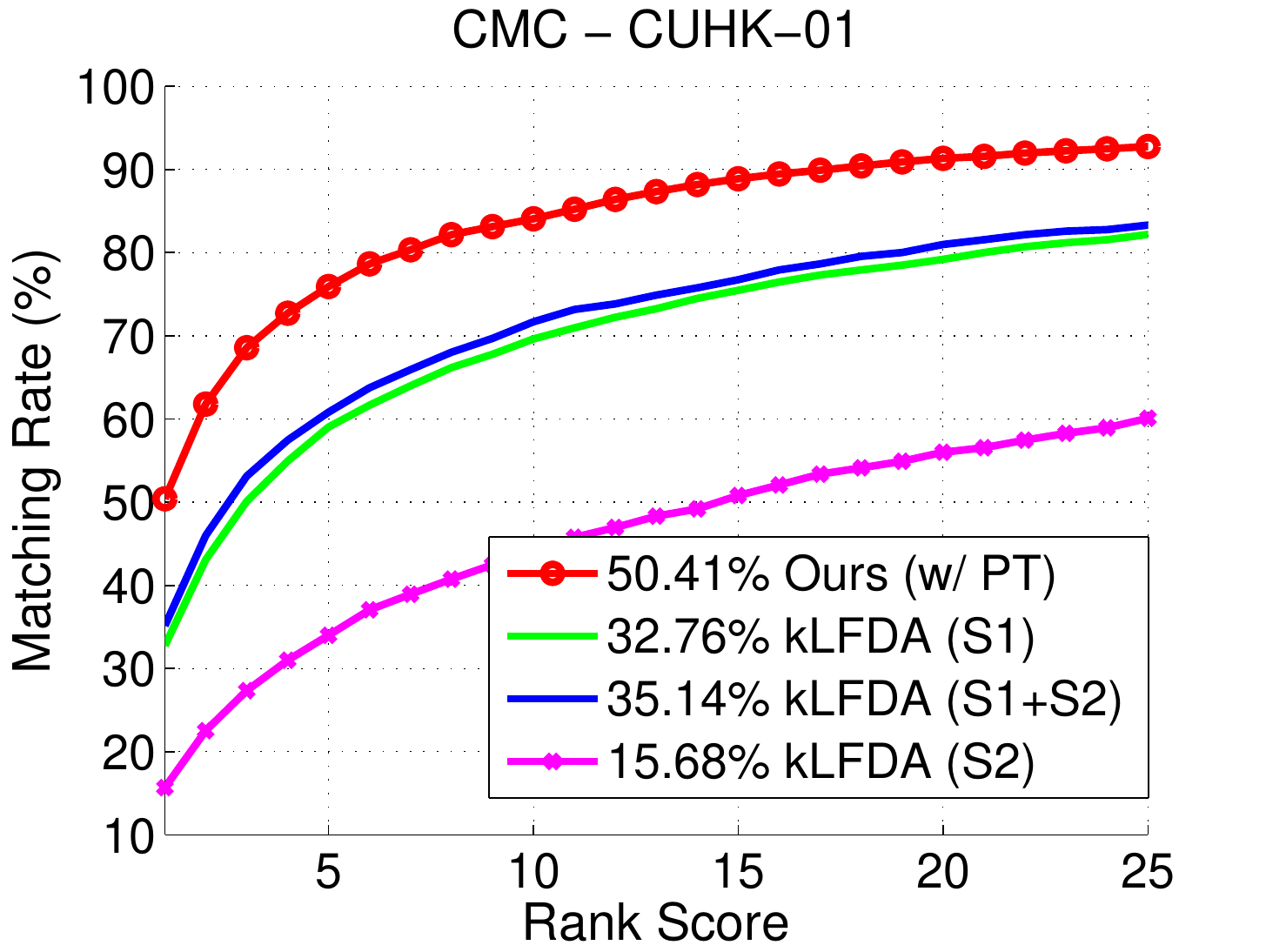}}
\caption{Comparisons of kLFDA performance with different training data on the VIPeR and CUHK-01 datasets. $\mathcal{S}_1$ denotes the original subset for training, and $\mathcal{S}_2$ denotes the data of P2-P5 from the CUHK-02 dataset. The CMC curves of our method that pre-trains the deep network with $\mathcal{S}_2$ are also presented for comparison.}
\label{fig:aug_training_cmp}
\end{figure}

However, one issue remains: it seems that the pre-training stage gives the proposed method an ``unfair" advantage compared to other state-of-the-art approaches, which have been devised under a constraint of only using the dataset subset for training. To fairly verify the effectiveness of our method, we removed the pre-training stage, trained the network with only the training subset of CUHK-01, and then compared it with FPNN on the CUHK-01 dataset, as shown in Figure \ref{fig:cmp_fpnn_nopre}. The rank-1 matching rate declines from 70.94\% to 55.11\% when the pre-training stage is excluded. Even so, our method still achieves a greater than 27\% improvement over FPNN. It is clear that the remarkable performance of our model does not simply come from the advantage of pre-training, although pre-training contributes to better performance indeed. For traditional metric-learning-based algorithms, \cite{ma2013domain} has experimentally showed that learning metrics from outside data under different distributions leads to severe deterioration in performance. Here we also compare the experimental results when learning kLFDA metrics with different training data to gain insights into how traditional methods are influenced by outside training data. Figure \ref{fig:aug_training_cmp} shows the comparisons on the VIPeR and CUHK-01 datasets. Let $\mathcal{S}_1$ denote the original subset for training, and $\mathcal{S}_2$ denotes the data of P2-P5 from the CUHK-02 dataset. In other words, traditional metric learning algorithms learn the models with $\mathcal{S}_1$ that comes from the same dataset. $\mathcal{S}_2$ is exactly the training set that we used to pre-train the deep network. We evaluated three types of training sets: $\mathcal{S}_1$, $\mathcal{S}_2$, and $\mathcal{S}_1\cup\mathcal{S}_2$. It is demonstrated that (1)~learning the metrics from $\mathcal{S}_2$ leads to the worst performance, because there exist significant differences between the distributions of $\mathcal{S}_2$ and the test target sets; (2)~little performance gains are observed if we augmented the training set with outside data (i.e., using $\mathcal{S}_1\cup\mathcal{S}_2$), compared to that learned with $\mathcal{S}_1$; (3)~different from traditional metric learning approaches, our proposed method can better leverage the outside data to learn more discriminative representation that possesses strong generalization ability. That is the reason why our proposed method exploits the pre-training strategy.

\subsubsection{Analysis of ranking unit sampling} \label{sssec:ranking_unit}
\begin{figure}
  \centering
  \subfigure[]
  {\label{subfig:refsam_viper}
  \includegraphics[width=0.23\textwidth, angle=0]{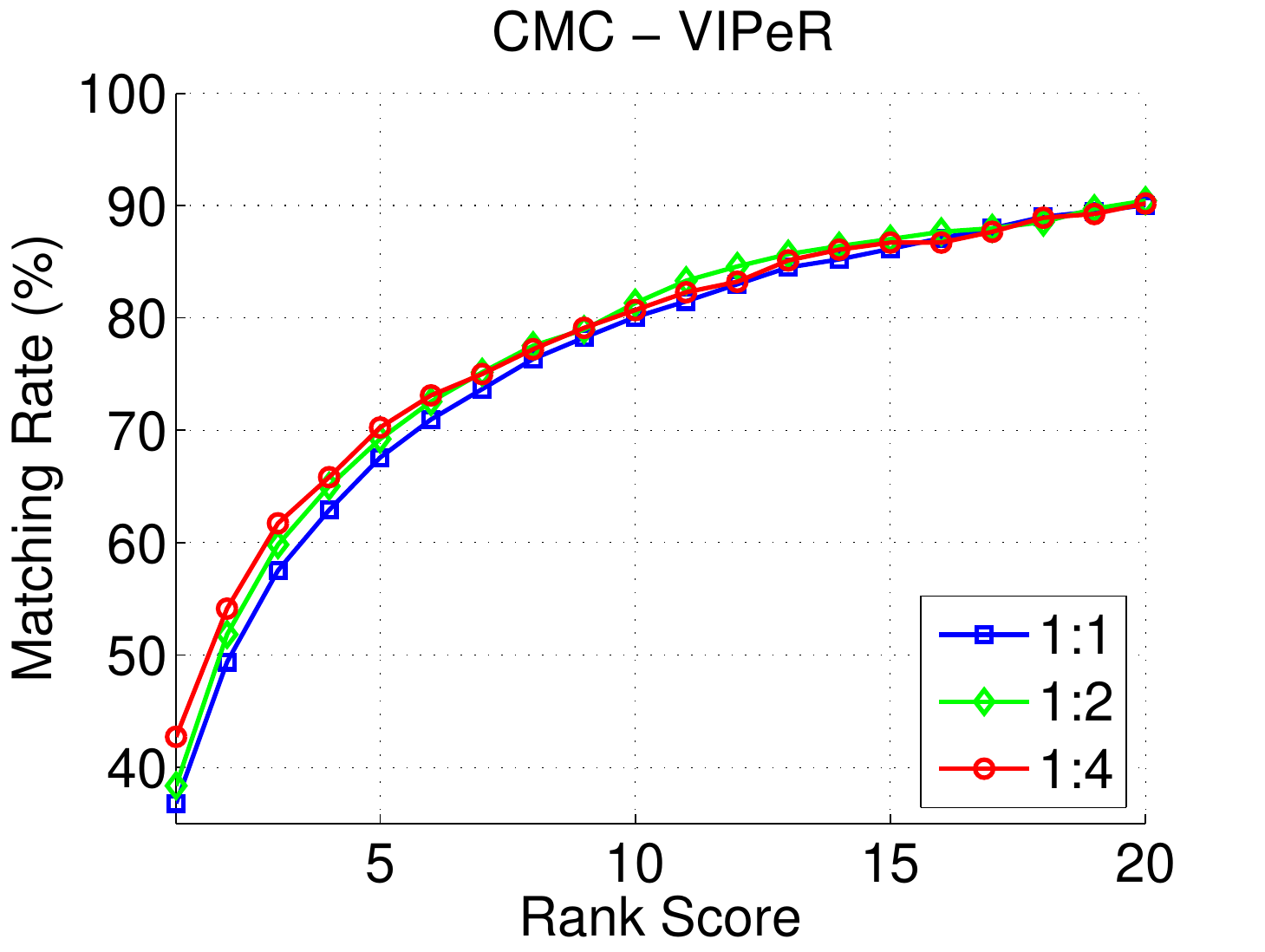}}
  \subfigure[]
  {\label{subfig:refsam_cuhk}
  \includegraphics[width=0.23\textwidth, angle=0]{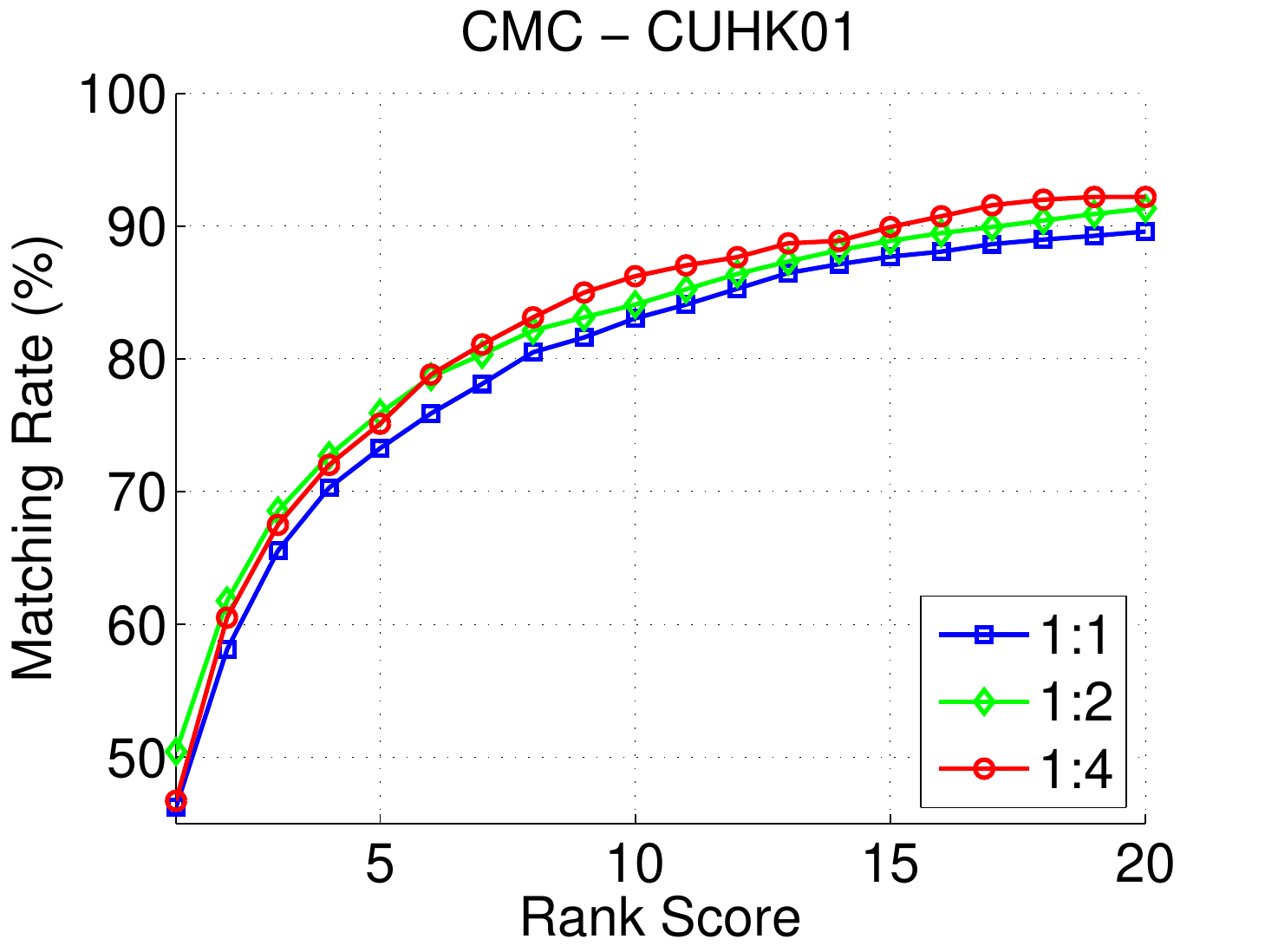}}
  \caption{Analysis of ranking unit sampling. The CMC curves on the VIPeR dataset (a) and CUHK-01 dataset (b) with different positive-negative ratios in the ranking units. Best viewed in color.}
  \label{fig:refsam}
\end{figure}

We here analyze the effectiveness of the ranking unit sampling method. Recall that we initially trained the deep network with $|\mathcal{R}_x|=1$, i.e., a positive-negative ratio (abbreviated as ``ratio" for convenience) of 1:1, and then gradually increased the $|\mathcal{R}_x|$ up to 4 (a ratio of 1:4). Figure \ref{fig:refsam} shows the CMC curves with different ratios of the reference sets in the ranking units on the VIPeR and CUHK-01 datasets. The CMC curves with a ratio of 1:2 consistently surpassed those with a ratio of 1:1 on both datasets, but no significant improvement was observed when we increased the ratio of the negative pairs to 4. On the VIPeR dataset, the CMC with a 1:4 ratio achieved a better rank-1 matching rate but performed nearly the same after rank-5. On the CUHK-01 dataset, the CMC with a 1:4 ratio worsened slightly but showed better performance after rank-7. We have also conducted more experiments on the change of ratio. The experimental results suggest that increasing $|\mathcal{R}_x|$ gives a small boost in re-identification performance, but the improvement with an increase in $|\mathcal{R}_x|$ is near saturation until 4. We conclude that the randomly sampled ranking units with a ratio of 1:2 can approximately replace the whole gallery set $\mathcal{G}$ during optimization. All of the results compared with the state-of-the-art algorithms had a ratio of 1:2.

\section{Conclusion} \label{sec:conclusion}
In this paper, we formulate the person re-identification task as a learning-to-rank problem, and propose a ranking model that learns a similarity metric that tends to place the true match of a probe at the top by penalizing ranking disorders in the gallery. A deep CNN is utilized to build a relation between image pairs and their similarities. These two components are then seamlessly integrated into a unified deep ranking framework that conducts similarity computing in one shot via joint representation learning directly from raw pixels without feature engineering. Extensive experimental results clearly demonstrate the effectiveness
of our proposed approach.

In the future, we would like to further explore ways to make full use of larger-scale outside data for network learning. In addition, we plan to explore how to adapt our approach to video data, i.e., how to measure the similarity of two sequences of detected pedestrian images. Finally, our framework can easily be applied to improve other learning-to-rank tasks such as relative attributes and image interestingness prediction.


%


\section*{Acknowledgment}
The authors deeply appreciate the anonymous reviewers for thoughtful comments that have improved this paper.

\ifCLASSOPTIONcaptionsoff
  \newpage
\fi



%

{
\bibliographystyle{IEEEtran}
\bibliography{egbib}
}

%







\end{document}